\documentclass[10pt,twocolumn,letterpaper]{article}

\usepackage[pagenumbers]{cvpr} 

\definecolor{cvprblue}{rgb}{0.21,0.49,0.74}
\usepackage[pagebackref,breaklinks,colorlinks,allcolors=cvprblue]{hyperref}
\usepackage[most]{tcolorbox}
\tcbset{
    colback=blue!5!white,
    colframe=blue!50!black,
    arc=2mm,
    boxrule=0.5pt,
    left=4pt,right=4pt,top=4pt,bottom=4pt,
}

\usepackage{amsthm}

\theoremstyle{plain} 

\theoremstyle{definition} 

\theoremstyle{remark} 

\def\paperID{} 
\def\confName{CVPR}
\def\confYear{2026}

\title{Towards Generalized Multi-Image Editing for Unified Multimodal Models}

\author{%
    Pengcheng Xu\textsuperscript{\rm 1,2}\footnotemark[3] \hspace{.1in}
    Peng Tang\textsuperscript{\rm 2}\footnotemark[2] \hspace{.1in}
    Donghao Luo\textsuperscript{\rm 2} \hspace{.1in}
    Xiaobin Hu\textsuperscript{\rm 2} \hspace{.1in} 
    Weichu Cui\textsuperscript{\rm 2} \hspace{.1in}
    Qingdong He\textsuperscript{\rm 2} \\
    Zhennan Chen\textsuperscript{\rm 3} \hspace{.1in}
    Jiangning Zhang\textsuperscript{\rm 2} \hspace{.1in} 
    Charles Ling\textsuperscript{\rm 1} \hspace{.1in} 
    Boyu Wang\textsuperscript{\rm 1}\footnotemark[1]  \vspace{.2cm} \\
    \textsuperscript{\rm 1}Western University  \hspace{.1in} 
    \textsuperscript{\rm 2}Tencent YouTu Lab \hspace{.1in}
    \textsuperscript{\rm 3}Nanjing University \\
    \small\href{https://github.com/Pengchengpcx/MIE-UMM}{Project Page: MIE-UMM}
    \vspace{-5mm}
}

\begin{document}
\maketitle
\let\thefootnote\relax\footnotetext{$^*$Corresponding author. $^{\dagger}$Project lead. $^{\ddagger}$Internship in Tencent.}

\begin{abstract}
Unified Multimodal Models (UMMs) integrate multimodal understanding and generation, yet they are limited to maintaining visual consistency and disambiguating visual cues when referencing details across multiple input images. 
In this work, we propose a scalable multi-image editing framework for UMMs that explicitly distinguishes image identities and generalizes to variable input counts. 
Algorithmically, we introduce two innovations: 1) The learnable latent separators explicitly differentiate each reference image in the latent space, enabling accurate and disentangled conditioning. 2) The sinusoidal index encoding assigns visual tokens from the same image a continuous sinusoidal index embedding, which provides explicit image identity while allowing generalization and extrapolation on a variable number of inputs. To facilitate training and evaluation, we establish a high-fidelity benchmark using an inverse dataset construction methodology to guarantee artifact-free, achievable outputs. Experiments show clear improvements in semantic consistency, visual fidelity, and cross-image integration over prior baselines on diverse multi-image editing tasks, validating our advantages on consistency and generalization ability.
\end{abstract}

\vspace{-3mm}
\section{Introduction}
Unified Multimodal Models (UMMs) have recently unified multimodal understanding and generation by integrating multimodal large-scale language models (MLLMs) with diffusion-based image generators~\cite{seed-x,emu2,emu3,qwen,dong2025seedvl}. Such hybrid systems can interpret complex visual-textual instructions and generate corresponding images. However, current editing methods based on the UMM~\cite{yu2025anyedit,wei2024omniedit,li2025superedit,zhao2024ultraedit,sheynin2024emuedit,huang2024smartedit,liu2025step1xeditpracticalframeworkgeneral,labs2025flux_kontext,liu2025step1x,wang2025seededit} mainly maintain \textit{semantic} alignment between inputs and outputs, while \textit{visual} consistency—preserving appearance, identity, and structure is mostly limited to one single image and begin deteriorating in multi-image editing, worse still, when extrapolating the exceeding number of input images in training data.

A broader paradigm for UMMs is that the output should maintain \textbf{both semantic and visual consistency to the variable-length multimodal input data, which genuinely unifies the understanding, generation, and editing in the multi-image setting}. This capability is fundamental for wide-ranging applications, including multi-subject ID generation~\cite{wu2025uno_flux,ma2024subject,chen2023photoverse,wang2025msdiffusion}, style transfer~\cite{wang2024instantid,wang2025omnistyle}, virtual try-on~\cite{feng2025omnitry,guo2025any2anytryon,han2018viton}, and advanced editing tasks that rely on referencing details across multiple source images~\cite{wu2025omnigen2,xia2025dreamomni2,qwen}. All these tasks share a common formulation: receive multiple reference images and textual instructions as input, and produce an output that is both semantically aligned with the textual guidance and visually consistent with the given image identities. 

However, current UMMs based on the MLLM–Diffusion hybrid architectures face a fundamental bottleneck:
When multiple reference images and text are provided, the model cannot effectively encode which latent feature from the VAE corresponds to which input image and generalize, which limits precisely referencing visual contents across multiple different images.

\begin{figure*}[t]
\centering
\includegraphics[width=0.95\linewidth]{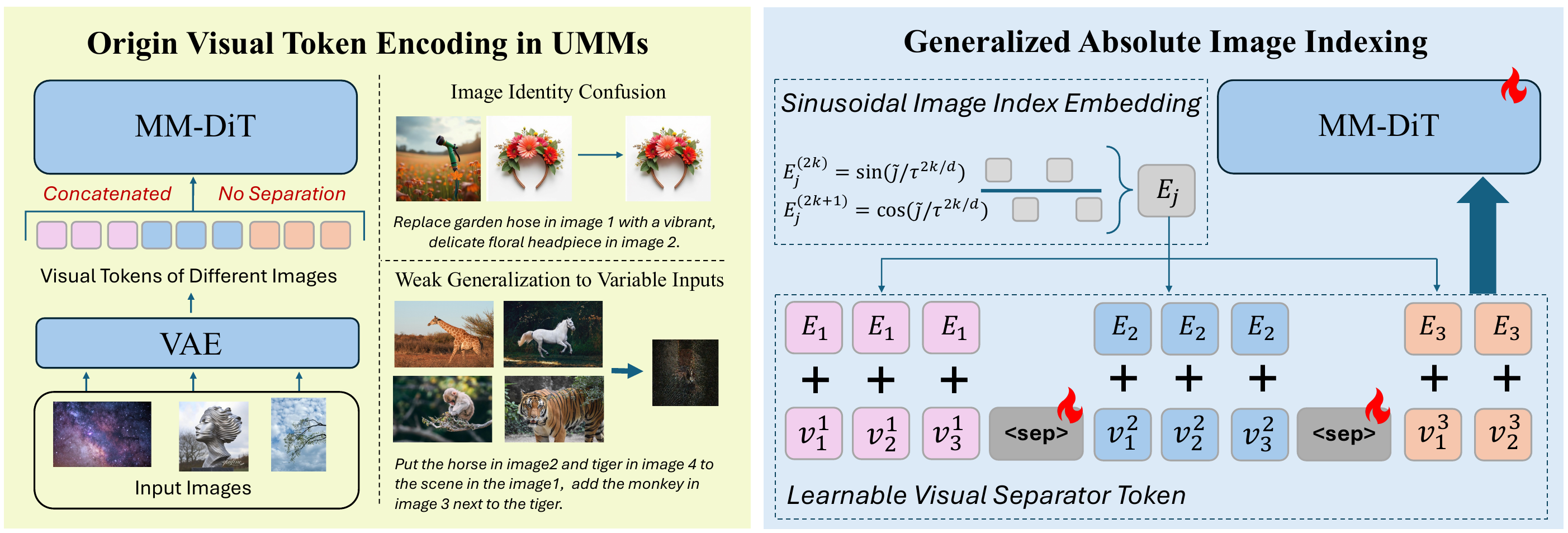}
\vspace{-3mm}
\caption{\textbf{Demonstration of the encoding of visual tokens behind the VAE in hybrid UMM and our design}. The arrangement of visual tokens lacks separation and the awareness of the image index. This can lead to confusion of instance identities, misinterpretation of the image index, and a lack of generation of an unseen number of input images.}
\label{fig:attn_analysis}
\vspace{-5mm}
\end{figure*}

Concretely, such deterioration is attributed to two reasons. First, the standard positional or rotary encodings (RoPE)~\cite{heo2024rotary} used in transformer backbones (e.g., MM-DiT)~\cite{dit} primarily capture \textit{relative ordering} between tokens but fail to preserve \textit{absolute positional} identity between images.
Consequently, as shown in Figure~\ref{fig:attn_analysis}, when multiple image latents are concatenated, the model tends to confuse instance identities, misinterpret the text’s image-specific references, and generate outputs that lose per-image consistency in both semantics and visuals. 
Especially when the resolution of images and the corresponding number of tokens vary, the relative distance modeling shows a deficiency in distinguishing images.

Second, the training of current unified models is often limited to a \textit{finite number} of image inputs. This rigid but practical situation raises challenges for generalization to a variable number of references, thereby restricting scalability and generalization in real-world scenarios. Thus, the lack of extrapolable index-awareness and disentangled latent composition are two obstacles to achieving generalizable multimodal generation and editing. 

To systematically resolve these bottlenecks, we first construct the multi-image editing dataset and evaluation benchmark, and then propose two innovations to enable the distinction of the visual tokens from different images. 
Specifically, on the dataset construction, we employ an inverse construction methodology to guarantee high fidelity and artifact-free outputs. 
By starting with a high-quality ground truth target image, we utilize an instruction-based editing model to reverse-engineer the input—specifically by simulating object addition, deletion, or replacement to obtain the necessary source and reference input images. This inverse process inherently ensures the desired output and substantially reduces artifacts, which mitigates learning the copy-and-paste issues and artifacts in generated output images.

On the algorithm part, our key insight is that achieving scalable visual consistency requires explicit image-wise separation and extrapolable index awareness in the latent space, beyond standard relative encoding. 
So, we first propose the index-aware latent separation, which introduces learnable separator tokens, serving as explicit boundaries between each image latent.
Unlike conventional RoPE-based positional encodings that only encode relative distances, these learnable separators allow the model to distinguish absolute image positions within the multimodal sequence, enabling accurate cross-image reference tracking. 
Second, we propose the sinusoidal index embedding to indicate the image index of tokens in the sequence. Specifically, it assigns the visual tokens from \textit{the same image} with the \textit{same} continuous index embedding.
Since the index embedding is built on the sinusoidal function, this also provides the extrapolation ability for different numbers of input images. 
Thus, these learnable separators and index embeddings explicitly inform the model which visual tokens belong to which image, akin to giving each reference a unique identity, effectively preventing pixel confusion across different reference images, thereby ensuring high-fidelity, visual consistency in the output. 

We summarize our contributions and findings as follows:

\begin{itemize}
    \item \textit{Algorithmic}: We propose a scalable multi-image editing framework with learnable visual separators and sinusoidal index embeddings that explicitly distinguish image identities and enable extrapolation to variable input numbers, achieving disentangled, identity-preserving multimodal representation.
    \item \textit{Dataset}: We build a high-fidelity multi-image editing dataset and benchmark via inverse dataset construction, ensuring artifact-free, achievable ground truths, providing a comprehensive evaluation on various editing types, scenarios, and numbers of input images.
    \item \textit{Empirical validation}: Experiments demonstrate that our method mitigates the cross-image confusion, enhances visual fidelity and consistency, and generalizes to unseen numbers of reference images.
\end{itemize}

\vspace{-2mm}
\section{Related Work}
\noindent\textbf{Unified Multimodal Models}. Recent unified multimodal models aim to unify multimodal understanding and image generation within a unified framework, enabling the understanding of complex multimodal instructions, and generating images more flexibly. There are mainly two categories of UMMs. The first is the hybrid UMM that assembles the MLLM for multimodal understanding and the diffusion model for image generation by training lightweight connectors or learnable tokens~\cite{metaquery, wu2025qwenimagetechnicalreport, chen2025blip3, labs2025flux_kontext, liu2025step1xeditpracticalframeworkgeneral, chen2025unireal, fu2025univg, chen2025multimodal}, which generally require less data and resources for training. The second is the native UMM that trains the multimodal understanding and image generation within a new and unified framework from scratch. Such a design aims to achieve better fusion of image and text modalities within a unified network as well as stronger collaborative understanding and generation abilities~\cite{chameleon, transfusion, xie2024show, xie2025show, janus2024, januspro2025, janusflow2024, emu3, tong2024metamorph, deng2025bagel}, but practically faces challenges in scaling and coordinating the training of generating text and images. Nevertheless, the current shared issue of these two categories is the generalization and scalability of visual consistency when referencing multiple images. The recent work Query-Kontext~\cite{song2025query} and DreamOmini2~\cite{xia2025dreamomni2} tackle the multi-image editing tasks by commonly shifting the RoPE to enlarge the relative distance of tokens from different images to avoid confusion. However, this lacks effectiveness in distinguishing image indexes and empirically does not generalize well to an extrapolated number of reference images.
We adopt the hybrid UMM as the backbone due to its large generation capacity and quality, and further propose explicitly separating and indexing tokens from different images, which shows better generalization on visual consistency when referencing multi-image inputs.

\noindent\textbf{Multi-image Generation and Editing}.  From a general perspective of multimodal image generation, many tasks share the same formulation that takes multiple images and textual instructions as input, and outputs an image conditioned on these input images with visual consistency. Concretely, the virtual try-on~\cite{xu2025ootdiffusion,guo2025any2anytryon,feng2025omnitry,jiang2024fitdit,han2018viton} accepts multiple images of garments, accessories, wearable objects, and a person to synthesize the person with all these try-ons. The multi-subject generation, such as UNO~\cite{wu2025uno_flux}, UMO~\cite{cheng2025umo}, and MultiCrafter~\cite{wu2025multicrafter}, accepts multiple reference images and a text instruction to compose a new scene while preserving the identities of the input images. Similarly, DreamO~\cite{mou2025dreamo} and USO~\cite{wu2025uso} accept the subject and style images for generating output resembling the input's identity and style simultaneously. Besides these, some advanced editing methods also take multiple images as input, either for visual reference~\cite{chen2024anydoor,li2024photomaker,zhou2024migc,chen2024zero} or for understanding complex multimodal instructions~\cite{huang2024smartedit,liu2025step1x,fu2025univg,wang2025seededit,wu2025qwen}. However, these frameworks mostly focus on one specific sub-task but cannot solve all as a unified model. Thus, to gain a general UMM that can treat all these tasks with the same formulation and solve them within a unified model, it is crucial to maintain the visual consistency and generalize when referencing multiple and different images. Our research proposes strategies to maintain visual consistency in multi-image scenarios, and can adapt to different generation and editing tasks within a UMM.

\section{Method}
We aim to equip a UMM with generalization ability to preserve visual consistency across multiple reference images. Section~\ref{sec:method_pre} briefly reviews the hybrid UMM architecture based on MLLM-Diffusion and widely adopted multimodal RoPE~\cite{heo2024rotary,wu2025qwenimagetechnicalreport}. Section~\ref{sec:image_indexing} then describes how to enable the UMM distinguishes tokens from different images to enable correct cross-image reasoning and referencing (e.g., adding an object from image 1 to image 2).

\subsection{Multi-Image Visual Token Encoding in UMM}
\label{sec:method_pre}
Generally, a hybrid UMM~\cite{xia2025dreamomni2, wu2025omnigen2} combines an MLLM~\cite{qwen} with a diffusion transformer (e.g, MM-DiT)~\cite{wu2025qwen, esser2024scaling, li2024hunyuan}, and thus uses two image encoders. A \textit{semantic} encoder (e.g., SigLip~\cite{zhai2023sigmoid}) provides image semantics to the MLLM, while a \textit{visual} encoder (e.g., VAE) extracts pixel-level features that govern visual consistency between the multiple input images and the edited output. As shown in Figure~\ref{fig:attn_analysis}, before entering MM-DiT, tokens from different images are reshaped and concatenated along the height and width dimensions (Eq.~\ref{eq:vtoken}). Let $v_i^j \in \mathbb{R}^{1 \times HW \times C}$ denote token $i$ from image $j$, where $HW$ flattens spatial dimensions and $C$ is the channel size. This concatenation does not explicitly mark image identity; instead, the model relies on RoPE to capture relative token distances and implicitly separate tokens from different images, as discussed next.
\begin{equation}
[\mathrm{~}v_{1}^1,v_{2}^1,v_{3}^1,v_{4}^1,\ldots,v_{1}^2,v_{2}^2,v_{3}^2,v_{4}^2,\ldots,v_{1}^j,v_{2}^j,v_{3}^j,\ldots]
\label{eq:vtoken}
\end{equation}

\noindent\textbf{Multimodal Rotary Position Embedding}. The multimodal RoPE is a three-dimensional multimodal system, covering the frame, height, and width dimensions~\cite{qwen}. To encode the local spatial layouts and global inter-image relationships of multiple images and text, each input image (or frame) $I_j$ is first tokenized into a 3D grid of shape $(F_j, H_j, W_j)$, where $ F_j$ is the frame count (typically 1 for static images). The RoPE is calculated based on the image shapes. All image shapes are concatenated along the frame axis, forming a unified sequence of tokens:
\begin{equation}
  \mathcal{V}=[V_1,V_2,\ldots,V_N],\quad V_j\in\mathbb{R}^{F_j\times H_j\times W_j}
\end{equation}

This effectively treats multiple images as a pseudo-video~\cite{wu2025qwenimagetechnicalreport, xia2025dreamomni2}, assigning each image a unique frame index while preserving its 2D spatial layout. For each image $I_j$ with shape $(F_j, H_j, W_j)$, the model constructs frequency tables for these axes:
\begin{equation}
    \mathrm{pos_{freqs},neg_{freqs}}=f_{\mathrm{rope}}(\mathrm{axes_{dim}}=[F,H,W])
\end{equation}

After getting the frequency table, the frequency of each axis for an image token is calculated by the frame index $j$ and spatial location $h$ and $w$, and the frequency of this token is the concatenation of these three kinds of frequencies in Eq.~\ref{eq:freq}, and the final multimodal RoPE of each token $x_{j,h,w}$ is computed in Eq.~\ref{eq:rope}
\begin{equation}
\label{eq:freq}
f(j,h,w)=[\mathrm{~}f_\mathrm{frame}(j),f_\mathrm{height}(h),f_\mathrm{width}(w)\mathrm{~}]
\end{equation}
\begin{equation}
\label{eq:rope}
\mathrm{RoPE}(x_{j,h,w})=x_{\mathrm{even}}\cos(f(j,h,w))+x_{\mathrm{odd}}\sin(f(j,h,w))
\end{equation}
where $x_{\mathrm{even}}$ are the even-indexed dimensions of channel while $x_{\mathrm{odd}}$ are odd-indexed dimensions of channel. With this multimodal RoPE strategy, the model can capture relative angular distances between tokens across all three axes by $\Delta=(j_2-j_1,\mathrm{~}h_2-h_1,\mathrm{~}w_2-w_1)$, which aims to represent both local spatial structures between 2D patches and global inter-image ordering across different images.

\noindent\textbf{Limitations}. Although multimodal RoPE aims to capture both intra-image spatial and inter-image ordering distance, we empirically find two deficiencies of this mechanism: First, it mostly captures the \textbf{relative} position information but lacks the notion of \textbf{ absolute} image identity. Consequently, this makes the model deficient in inferring explicit image boundaries or stable reference identities, especially when reasoning about cross-image composition. Thus, the model may confuse the referenced image in the text instruction and output a reference image as shown previously in Figure~\ref{fig:attn_analysis}. Second, the multimodal RoPE does not generalize well when the number of input images exceeds the number of images in the training data, which does not benefit the scalability and generalization. 

\subsection{Generalized Absolute Image Indexing}
\label{sec:image_indexing}
\noindent\textbf{Motivation}. Since the standard multimodal RoPE models relative spatial relationships well but do not effectively encode absolute image index, nor do they clearly distinguish which image a token belongs to, we aim to augment RoPE with explicit identity and boundary cues to ensure that the Transformer distinguishes between different image contexts while retaining spatial precision. We present the following strategies to achieve the generalized and extrapolatable image index encoding.

\noindent\textbf{Learnable Visual Separator Token}. We introduce a learnable visual separator token $<\mathrm{sep>}\mathrm{~} \in \mathbb{R}^{1\times d \times C}$, inserted between the visual token sequences of consecutive images as follows,
\begin{equation}
[v_{1}^1,v_{2}^1,v_{3}^1,<\mathrm{sep>}\mathrm{~},v_{1}^2,v_{2}^2,v_{3}^2,<\mathrm{sep>}\mathrm{~},\ldots]
\label{eq:sep_vtoken}
\end{equation}

Similarly to Eq.~\ref{eq:vtoken}, the whole sequence is flattened, and the token is reshaped. $d$ is the width of the separator, which determines the number of learnable parameters. Note that we only insert this separator among the image tokens while the text tokens are unchanged. For implementation at the code level, we include the shared learnable token as part of the DiT, which is shared across all images. During training, it is updated through backpropagation with the standard flow matching loss.

The shared token $<\mathrm{sep>}\mathrm{~}$ is learnable, acting as a soft boundary that separates visual token groups to prevent feature mixing, and provides transition semantics between consecutive images. Unlike fixed delimiters, this learnable separator dynamically encodes the degree of interaction between adjacent image segments—enabling the model to modulate cross-image attention during multi-image editing and composition.

\noindent\textbf{Generalized Sinusoidal Index Embedding}. To complement the separator token, we assign every image a continuous sinusoidal index embedding $E_i$ that provides explicit image identity while allowing extrapolation to unseen image counts. For each image index $j \in [1, N]$, we compute a normalized index $\tilde{j} = j / N$ and define the sinusoidal image index embedding with the sinusoidal base $\tau$:
\begin{align}
& E_j^{(2k)}=\sin(\tilde{j}/\tau^{2k/C})\\
& E_j^{(2k+1)}=\cos(\tilde{j}/\tau^{2k/C})\\
& k=0,\ldots,C/2-1
\end{align}
All visual tokens from the same image share the same embedding:
\begin{equation}
    \hat{v}_i^j = v_i^j + E_j
\end{equation}
Thus, $E_j$ encodes the absolute identity of image $j$, complementing multimodal RoPE’s relative encoding of $(j,h,w)$ coordinates. Because it is sinusoidal and non-learnable, this embedding smoothly extrapolates to arbitrary numbers of input images (e.g., training with 2–4 images, testing with 5–6).

\noindent\textbf{Unified Transformer Encoding}. The final multimodal sequence fed to the MM-DiT is denoted in Eq.~\ref{eq:final_vtoken}. In this sequence, the multimodal RoPE encodes the relative spatial relationships of visual tokens. The sinusoidal index embedding provides the absolute image index embedding. The separator tokens introduce cross-image boundaries and transitions. Together, these establish a hierarchical positional system that models the local spatial structure, global image identity, and the inter-image segmentation.
\begin{equation}
[\hat{v}_{1}^1,\hat{v}_{2}^1,\hat{v}_{3}^1,<\mathrm{sep>}\mathrm{~},\hat{v}_{1}^2,\hat{v}_{2}^2,\hat{v}_{3}^2,<\mathrm{sep>}\mathrm{~},\ldots]
\label{eq:final_vtoken}
\end{equation}

\begin{figure*}[t]
\centering
\includegraphics[width=0.95\linewidth]{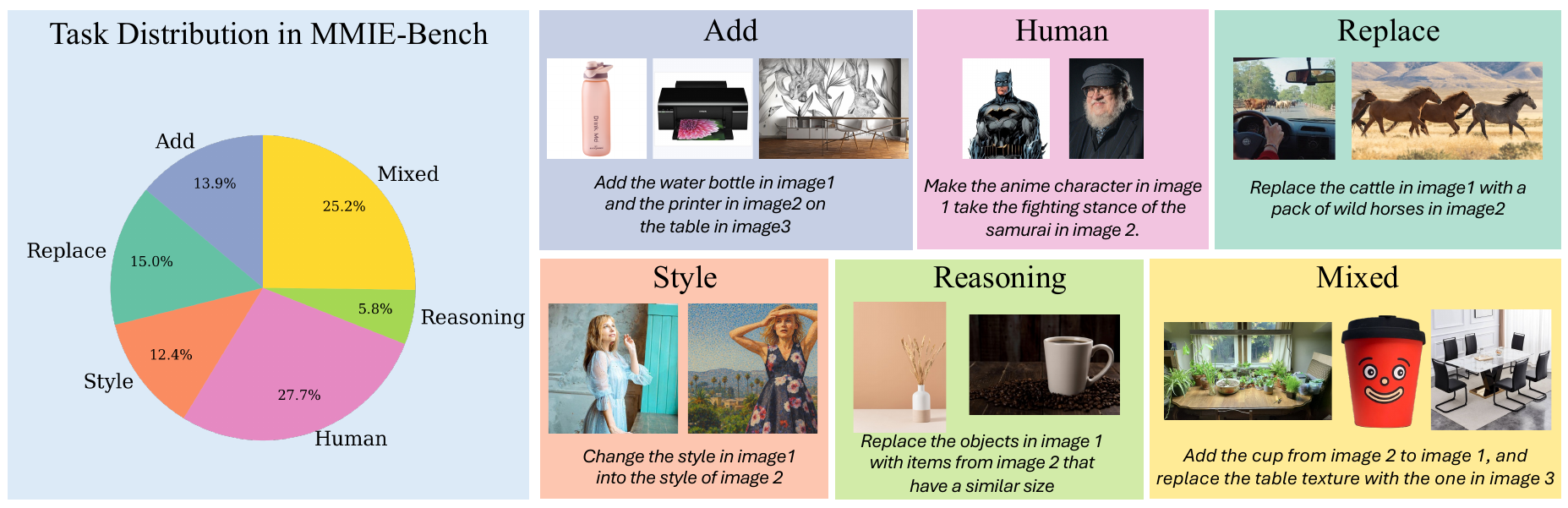}
\vspace{-3mm}
\caption{\textbf{Task distribution and editing examples of the MMIE-Bench}. The benchmark consists of six different editing tasks involving add, human, replace, style, reasoning, and mixed editing. These tasks also cover different objects, scenarios, and numbers of input images for comprehensive evaluation. All human portraits are from PIE and Echo-4o~\cite{ju2023direct, ye2025echo}.}
\label{fig:data}
\vspace{-5mm}
\end{figure*}

\section{Multi-image Editing Data Creation}
A central challenge in building a multi-image editing dataset lies in ensuring the visual fidelity of the edited results. Instead of synthesizing new targets through imperfect composition or blending, we adopt \textbf{a reverse, or inverse construction strategy}: we treat a naturally captured real image as the final edited result, and derive its corresponding input images backward. This inversion ensures that every edited result is photorealistic and contextually coherent, while the input images and textual instructions are systematically generated to simulate realistic multi-image editing.

\subsection{Source Data and Consistent Pair Mining}
For task related to object insertion and replacement, we start from two large-scale datasets, Subject200K~\cite{tan2024ominicontrol} and UNO1M~\cite{wu2025uno_flux}, each providing paired images depicting the same object under different environments. To guarantee subject alignment, we use an MLLM (e.g., Qwen2.5-VL 72B) to compute subject consistency scores for all candidate pairs and retain only high-consistency samples. This filtering removes ambiguous or cross-category pairs, yielding a clean and diverse set of semantically aligned object pairs suitable for editing simulation. For tasks related to style transfer, we select the Omnistyle-150k~\cite{wang2025omnistyle}, which consists of triplets of content, style reference, and stylized images as the source data. Similarly, we process and filter image triplets in which the stylized image exhibits good structural consistency with the content image and a consistent style with the reference image.

\subsection{Inverse Editing Synthesis}
\label{sec:data_cre}
We present the multi-image editing dataset and benchmark. For each consistent pair, we select the image with a richer and more complete background as the ground-truth edited result, and then derive the input image(s) by synthetically removing or replacing its key object. We leverage the Qwen-Edit~\cite{wu2025qwenimagetechnicalreport} to edit the single image. This reverse formulation naturally produces two editing types. For the addition tasks, we remove the shared subject from one image and use the complete image as the edited target, and use the other image as the reference for the deleted object. For the replacement task, we first use GPT-4o and the object list of the large-scale instance segmentation dataset to get the common objects to be replaced with. In this way, the constructed data can cover most daily used objects and benefit the generalization ability. Then, we replace the subject in one image with a randomly chosen object in our object list. Then, similarly, we use the other image as the reference for the replacement, and use the original first image as the edited target. For the style transfer task, since the dataset already consists of triplets, we only construct the editing instruction. In summary, this backward process ensures that all edited results are visually valid, real-world images rather than composite renderings. Please refer to the supplementary for details of the dataset.

\subsection{Multimodal Multi-Image Editing Benchmark}
To systematically evaluate the capabilities of multi-image editing models, we introduce the Multimodal Multi-Image Editing Benchmark (MMIE-Bench), a diverse and balanced testbed spanning six task categories: Addition (Add), Replacement (Replace), Style Transfer (Style), Human Editing (Human), Reasoning, and Mixed Add–Replace–Style (Mixed).
The benchmark contains 274 curated examples, each consisting of multiple input images, a textual editing instruction, and a final edited image. The number of input images varies from two to five. Figure~\ref{fig:data} illustrates the task distribution across the six categories. MMIE-Bench captures progressively complex editing scenarios:
\begin{itemize}
    \item Add / Replace / Style — focus on localized object or appearance transformations, and global style transfer.
    \item Human — emphasizes pose, expression, and clothing transfer across human or avatar subjects.
    \item Reasoning — requires abstract or in-context transformations beyond explicit instruction.
    \item Mixed — combines addition, replacement, and style cues in 3–4 image settings to test compositional reasoning.
\end{itemize}

We use the MLLM to evaluate models using three complementary dimensions:
1. Semantic Consistency: faithfulness to the instruction semantics. 2. Visual Fidelity: realism and absence of artifacts. 3. Multi-Image Integration: spatial and contextual coherence across sources. Each score ranges from 1–5 and is averaged to yield the final benchmark metric. MMIE-Bench thus provides a unified and fine-grained evaluation framework for scalable multi-image editing under multimodal understanding.

\begin{table*}[t]
\centering
\tabcolsep=0.1cm
\caption{
\textbf{Quantitative results on MMIE-Bench evaluated by two MLLMs}.
Left: results scored by Qwen2.5-VL(72B); Right: results scored by Doubao-1.6. Our method achieves consistent improvements under both evaluators across all task families.
}
\vspace{-3mm}
\label{tab:main_results}
\resizebox{\textwidth}{!}{
\begin{tabular}{lccccccc|ccccccc}
\toprule
 & \multicolumn{7}{c|}{\textbf{Qwen2.5-VL Evaluation}} & \multicolumn{7}{c}{\textbf{Doubao-1.6 Evaluation}} \\
\cmidrule(lr){2-8} \cmidrule(lr){9-15}
Method & Add & Replace & Style & Human & Reason & Mixed & Avg 
& Add & Replace & Style & Human & Reason & Mixed & Avg \\
\midrule
Qwen-Edit & 2.99 & 3.00 & 2.56 & 2.72 & 2.75 & 2.67 & 2.77 
& 3.66 & 3.35 & 2.45 & 2.79 & 2.83 & 2.63 & 2.95 \\
DreamOmni2 & 3.23 & 3.35 & 2.93 & 2.97 & 2.83 & 2.93 & 3.04 
& 3.92 & 3.94 & 3.21 & 3.23 & 3.14 & 3.22 & 3.44 \\
OmniGen2 & 3.26 & 2.82 & 2.93 & 3.07 & 2.79 & 3.09 & 3.03 
& 3.68 & 3.08 & 3.17 & 3.29 & 2.65 & 3.32 & 3.20 \\
\textbf{Ours} & \textbf{3.77} & \textbf{3.51} & \textbf{3.09} & \textbf{3.22} & \textbf{3.12} & \textbf{3.30} & \textbf{3.34} 
& \textbf{4.46} & \textbf{4.10} & \textbf{3.28} & \textbf{3.59} & \textbf{3.15} & \textbf{3.65} & \textbf{3.70} \\
\bottomrule
\end{tabular}
}
\vspace{-4mm}
\end{table*}

\begin{figure*}[t]
    \centering
    \includegraphics[width=0.16\textwidth]{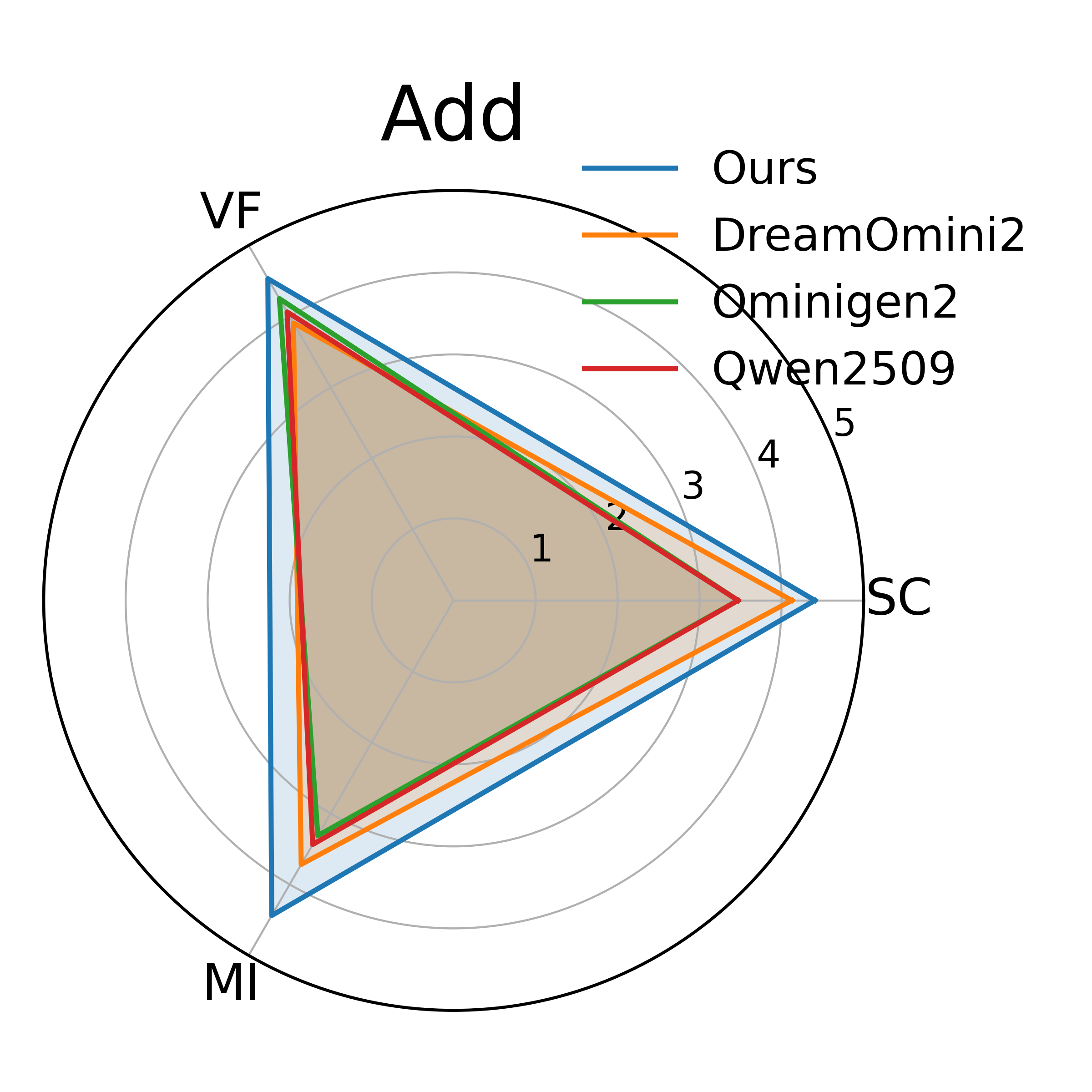}
    \includegraphics[width=0.16\textwidth]{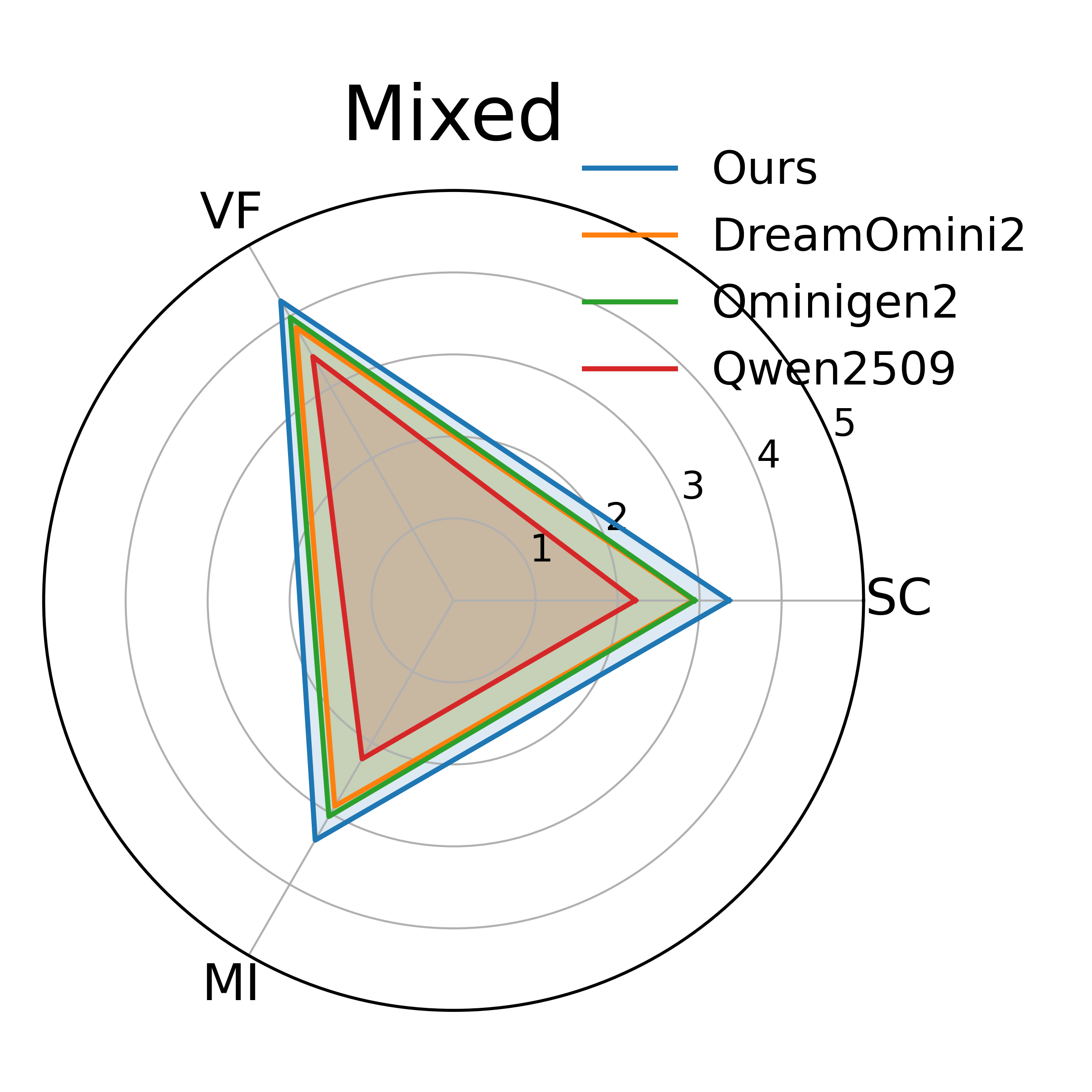}
    \includegraphics[width=0.16\textwidth]{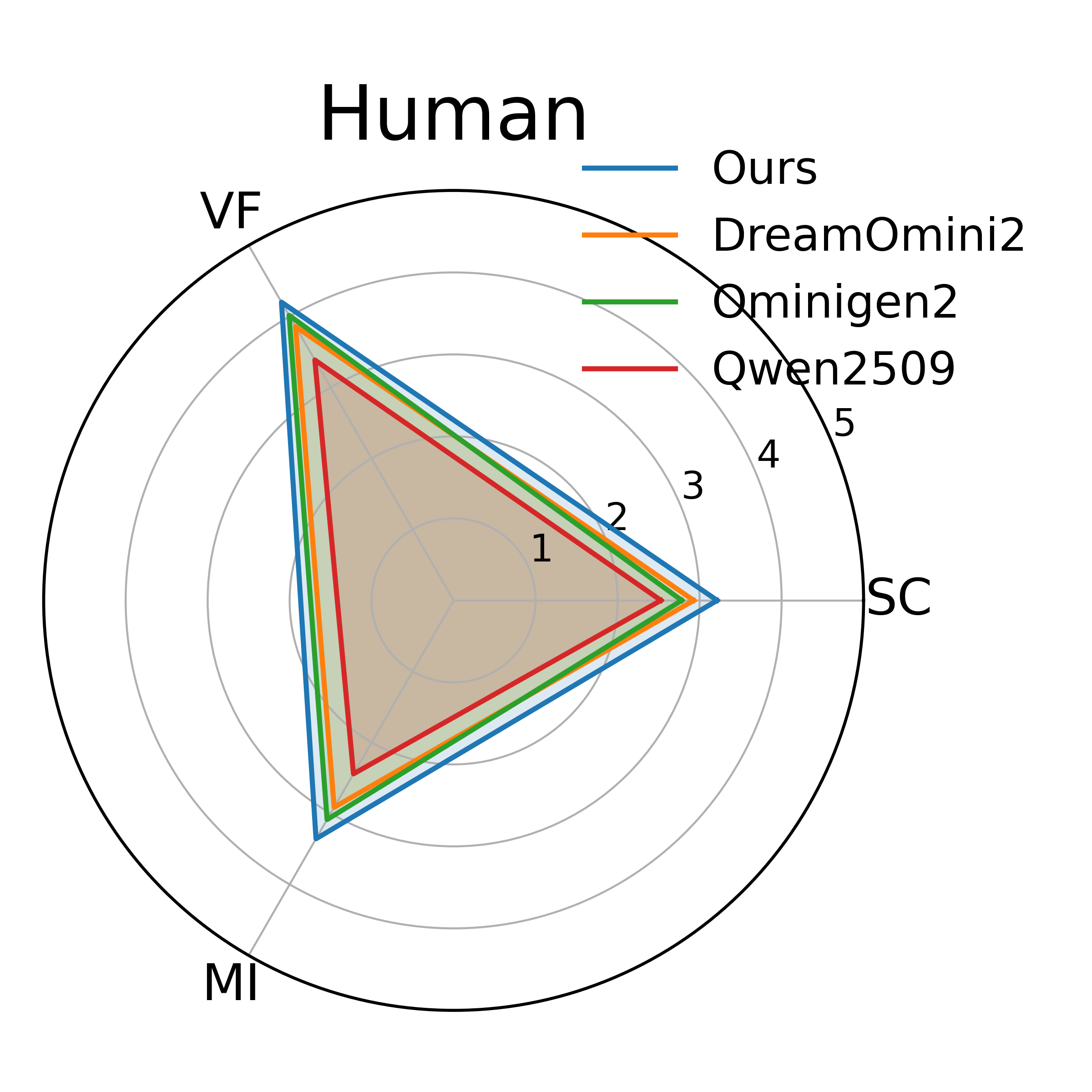}
    \includegraphics[width=0.16\textwidth]{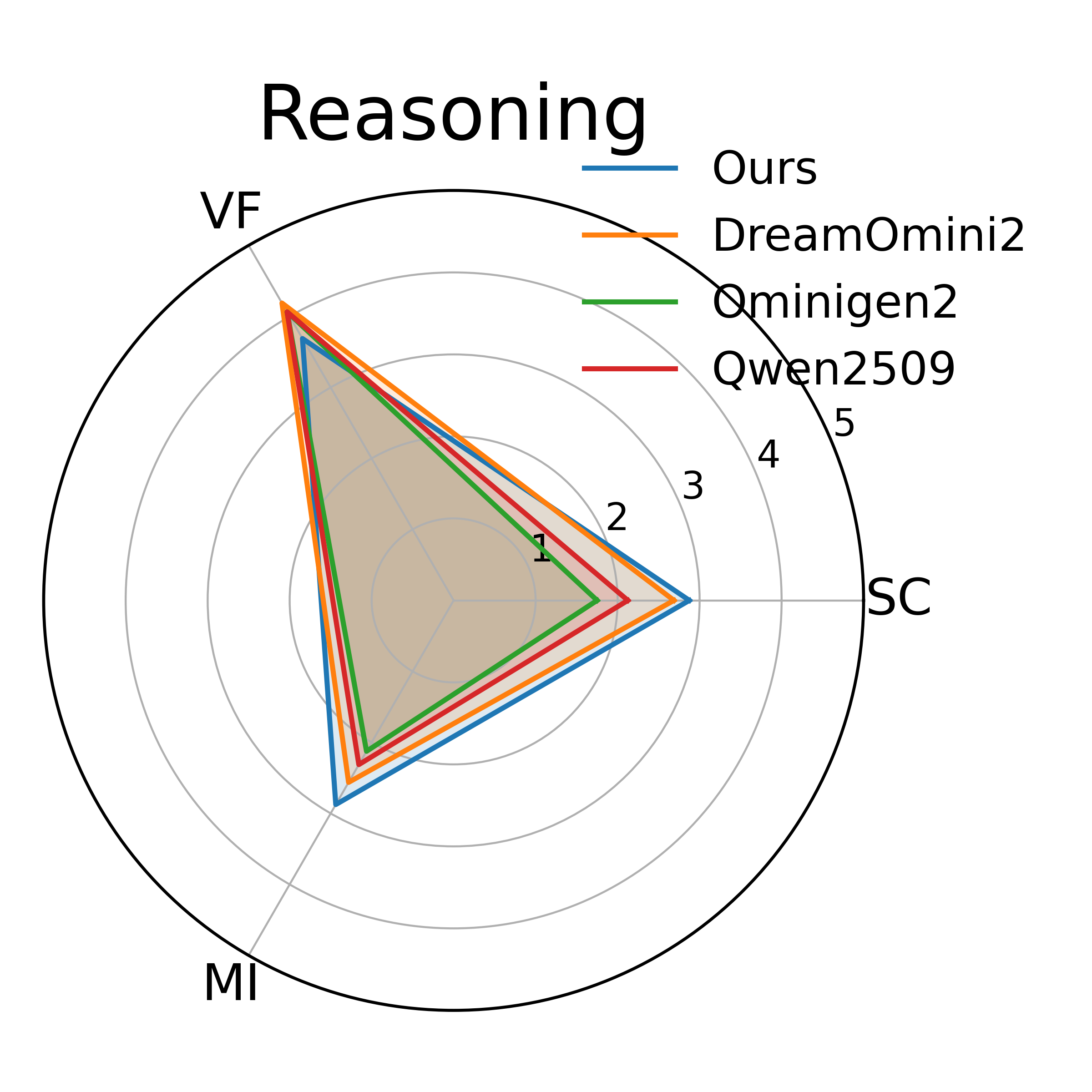}
    \includegraphics[width=0.16\textwidth]{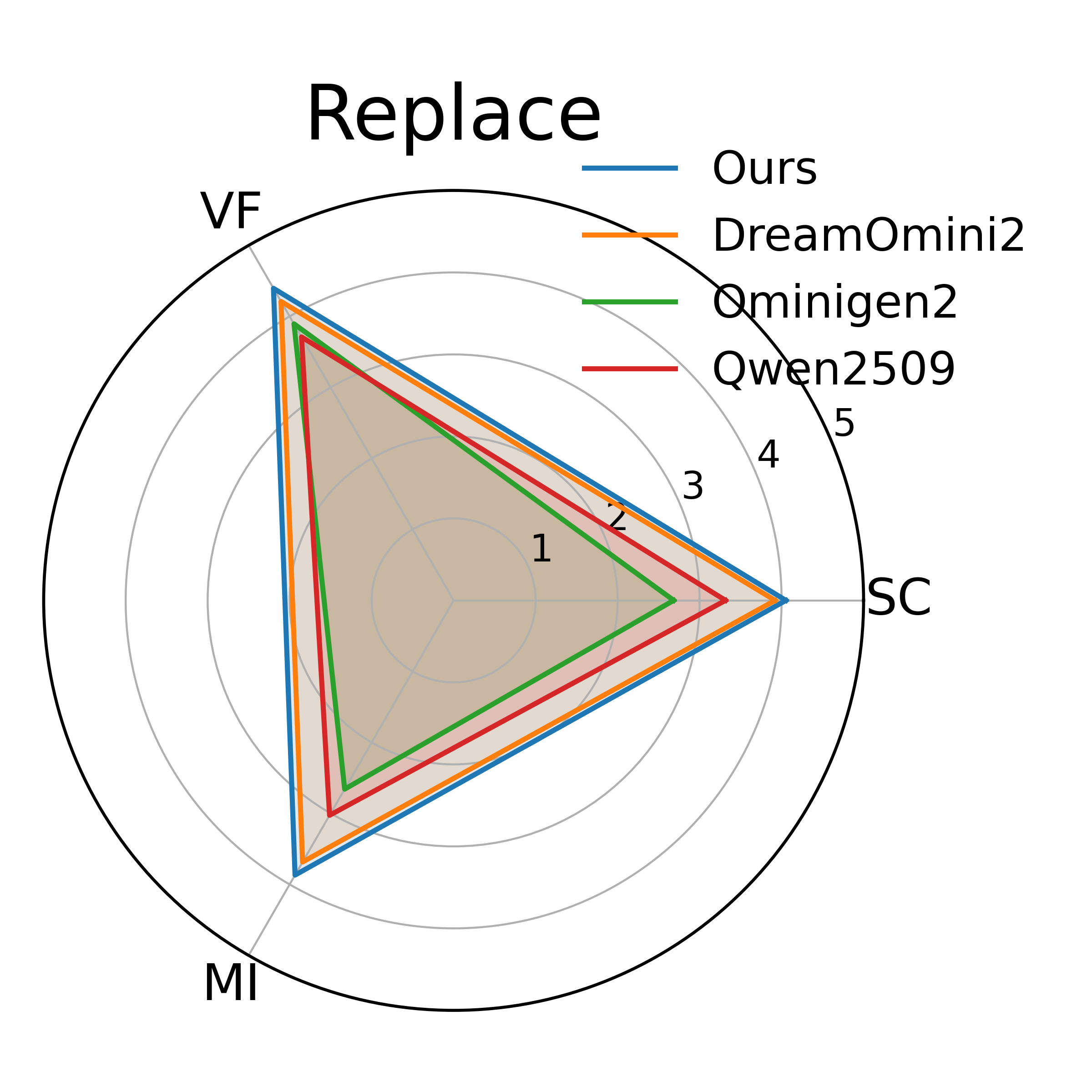}
    \includegraphics[width=0.16\textwidth]{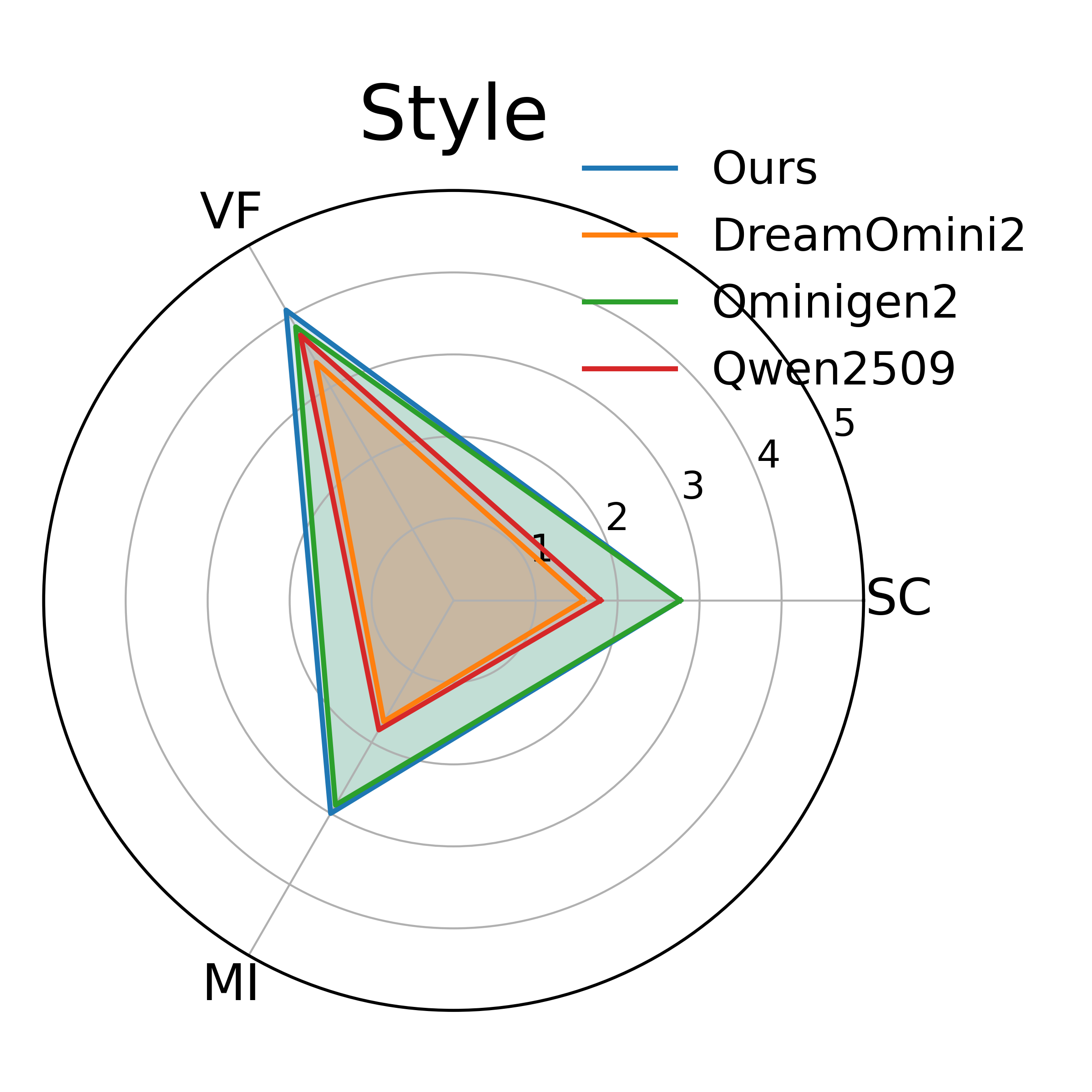}
    \vspace{-5mm}
    \caption{
        \textbf{Radar evaluation across six multi-image editing tasks by Doubao-1.6}. 
        Each radar chart compares four models over the three metrics: 
        Semantic Consistency (SC), Visual Fidelity (VF), and Multi-image Integration (MI). The metric score is rated from 1 to 5.
    }
    \label{fig:six_radar_row}
\vspace{-4mm}
\end{figure*}

\section{Experiments}

\subsection{Setup}
\noindent\textbf{Baselines and Implementations}. We compared three methods that support multimodal multi-image editing. All these three methods adopt the hybrid MLLM-Diffusion architecture to achieve the multi-image editing. The Ominigen2~\cite{wu2025omnigen2} unify the Qwen2.5-VL (3B)~\cite{qwen2vl} with  Lumina-Image 2.0~\cite{qin2025lumina} for diverse text-to-image and image editing tasks. The Qwen-Edit-2509~\cite{wu2025qwenimagetechnicalreport} retrain the Qwen2.5-VL (7B) with the MM-DiT with multi-image editing data. The DreamOmni2~\cite{xia2025dreamomni2} also adopts the Qwen2.5-VL (7B) as the MLLM and train the editing and generation models using LoRA on Flux Kontext~\cite{labs2025flux_kontext} to perform multimodal instruction-based editing and generation. For all comparison methods, we follow their official implementations for evaluation. For our method, we use the Qwen-Edit-2509 as our backbone, set the sampling steps as 40, and fix the output image resolution to 1328x1024 for all experiments. The classifier-guidance scale is set as 4.0. We use the same random seed in all experiments.

\noindent\textbf{Evaluation Data and Metrics}. We evaluate all models based on our proposed MMIE-Bench from three complementary metrics. The semantic consistency (SC) measures how accurately the output matches the instruction semantics. The visual fidelity (VF) evaluates perceptual realism and the absence of artifacts. The multi-image integration (MI) evaluates spatial and semantic coherence among multiple sources. We leverage multimodal understanding abilities of the MLLM to compare the input and output. For fair comparison, we use two different MLLMs, Qwen2.5-VL (72B) and Doubao-1.6. We write the prompt template to require the MLLMs to evaluate these three metrics. Please see the supplementary for concrete prompt templates. Each metric is rated in [1$\sim$ 5] and averaged to obtain the final score.

\subsection{Comparison with Previous Methods}
\begin{figure*}[t]
\centering
\includegraphics[width=1.0\linewidth]{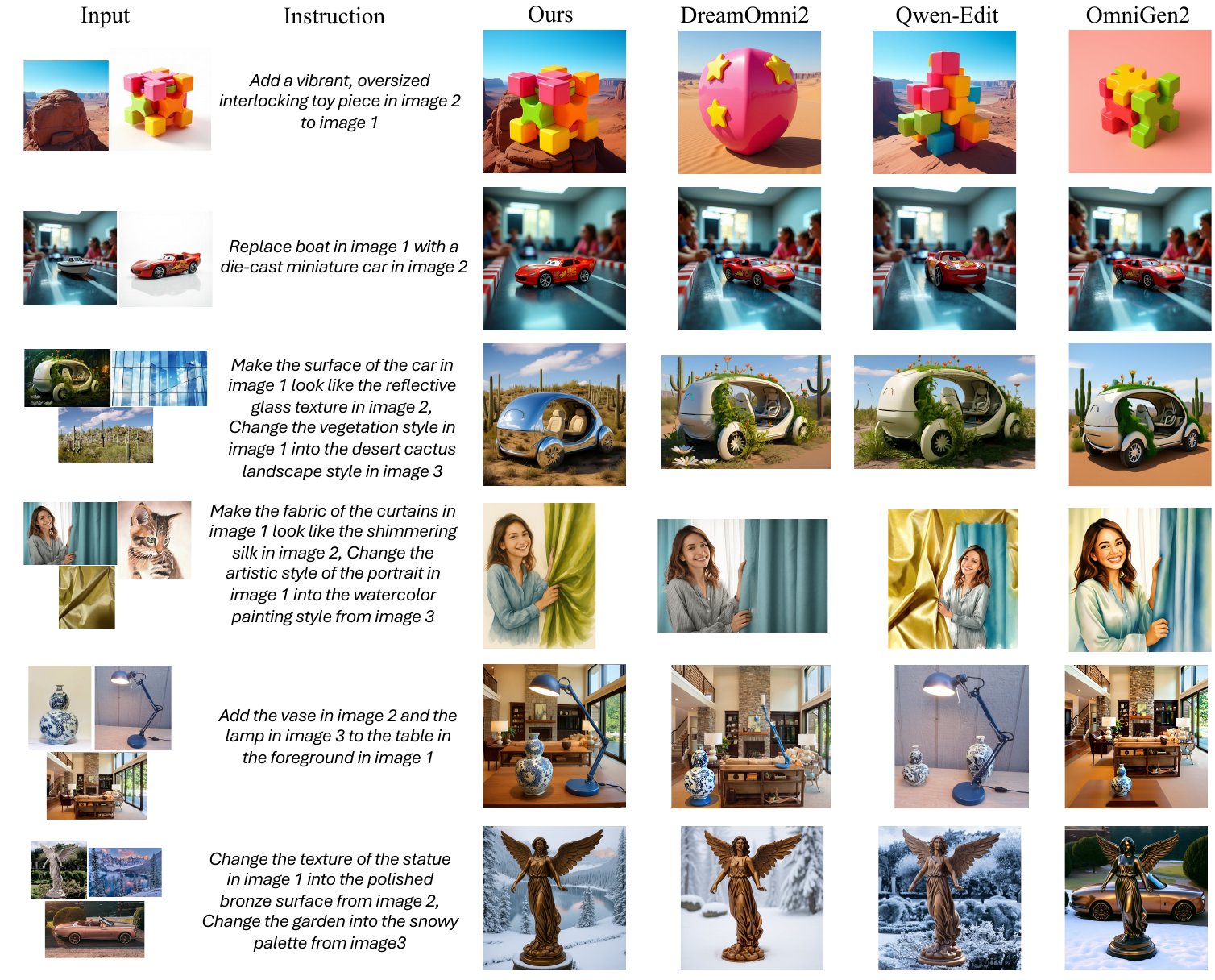}
\vspace{-5mm}
\caption{\textbf{Qualitative comparison on representative MMIE-Bench tasks}. Our method produces geometrically aligned, instruction-consistent, and compositionally coherent results across addition, replacement, texture transfer, and multi-style fusion tasks. All human data is from Echo-4o and PIE~\cite{ye2025echo,ju2023direct}.}
\label{fig:qualitative_results}
\vspace{-3mm}
\end{figure*}

We evaluate our method both quantitatively and qualitatively. Table~\ref{tab:main_results} summarizes the main comparison across six task families. Figure~\ref{fig:qualitative_results} demonstrates the qualitative results of different tasks, numbers of images, and scenarios. Our method achieves the best results across all categories, with strong gains on \textit{Mixed} task, which demands accurate multimodal understanding, distinction, and visual consistency on different images.

\noindent\textbf{Improvement over Baselines}. Compared to the original Qwen-Edit-2509, our methods improve the baseline performance by around 0.5 on all six tasks. This validates that our effectiveness in improving the distinction of visual tokens of different images clearly benefits the multi-image editing.

\noindent\textbf{Advantages on Complex and Mixed editing}. The Mixed task requires conducting add, replace, and style editing together. Our method outperforms the baseline Qwen-Edit-2509 by 0.37 on Reason and 0.63 on Mixed (based on Qwen). Based on Doubao-1.6, we get a 1.02 gain on Mixed. We also clearly outperform the second performance by 0.21 (on Qwen) and 0.33 (on Doubao) on Mixed. We attribute this to our advantages in distinguishing and manipulating cross-image visual contents, which then facilitates accurate reasoning and operation. 

\noindent\textbf{Multidimensional Evaluation}. Figure~\ref{fig:six_radar_row} shows the performance on three editing metrics over six tasks. Our method outperforms other methods on three metrics on most tasks. Especially for the Multi-image Integration (MI), our method shows clear advantages, which validate our advantages on the distinction of image identity, cross-image consistency, and compositional alignment.

\begin{table}[t]
\centering
\tabcolsep=0.12cm
\caption{
\textbf{User study on sampled MMIE-Bench}. Each task is evaluated based on 5 randomly sampled editing cases.
}
\vspace{-3mm}
\label{tab:user}
\resizebox{\linewidth}{!}{
\begin{tabular}{lccccccc}
\toprule
 & \multicolumn{7}{c}{\textbf{Qwen2.5-VL Evaluation}} \\
\cmidrule(lr){2-8}
Method & Add & Replace & Style & Human & Reason & Mixed & Avg \\
\midrule
Qwen-Edit      & 3.70 & 3.31 & 2.23 & 2.75 & 2.86 & 2.57 & 2.90 \\
DreamOmni2    & 3.13 & 3.58 & \textbf{4.10} & 3.18 & 3.12 & 3.11 & 3.37 \\
OmniGen2       & 3.40 & 3.41 & 2.71 & 2.90 & 2.92 & 3.08 & 3.07 \\
\textbf{Ours}  & \textbf{4.22} & \textbf{3.82} & 3.84 & \textbf{3.22} & \textbf{3.18} & \textbf{3.31} & \textbf{3.60} \\
\bottomrule
\end{tabular}
}
\vspace{-3mm}
\end{table}

\noindent\textbf{User Study}. We conduct the user study to verify the consistency between the MLLM evaluation and human preference. Similar to Table~\ref{tab:main_results}, we ask the user to rate each editing result based on three metrics of SC, VF, and MI. Table~\ref{tab:user} shows that the overall human preference is close to the MLLM evaluation results, and our method is favored on most tasks. See the supplementary material for details.

\subsection{Qualitative Results and Discussion}

Figure~\ref{fig:qualitative_results} shows representative examples across diverse multi-image editing scenarios, including object addition, texture transfer, style fusion, and object replacement. Prior multimodal editors frequently exhibit identity entanglement, partial modality transfer, and cross-image feature interference, revealing a structural limitation in architectures that rely purely on relative positional encodings. These failure modes manifest as inconsistent object boundaries, incomplete material propagation, and erroneous style dominance when multiple reference signals compete.

\noindent\textbf{Multi-image Object Addition}. Prior methods exhibit structural distortion and spatial drift, either failing to preserve identity or shifting locations. Our method accurately locates the objects and preserves the structural and fine appearance of the reference, demonstrating strong cross-image spatial reasoning.

\noindent\textbf{Cross-image Texture and Style transfer}. In tasks involving two or more style sources, baselines often transfer only partial material attributes or distort the original structure (e.g., glass → car body, silk → curtain). Our model faithfully maps the reference texture while retaining the original shape without leaking irrelevant visual cues from non-target references (e.g., bronze surface of car → statue).

\noindent\textbf{Object Replacement with Fine-grained Alignment}. In replacement scenarios, baseline methods do not preserve the identities well and generate with mild imagination, and the replaced object is not well aligned with the reference (e.g., the boat  → miniature car). Our approach delivers better integration of the reference object’s geometry.

Overall, these qualitative trends align with our motivation to improve cross-image referencing, alignment, and visual consistency in multi-image editing.

\section{Ablation and Analysis}

\begin{figure}[t]
\centering
\includegraphics[width=1.0\linewidth]{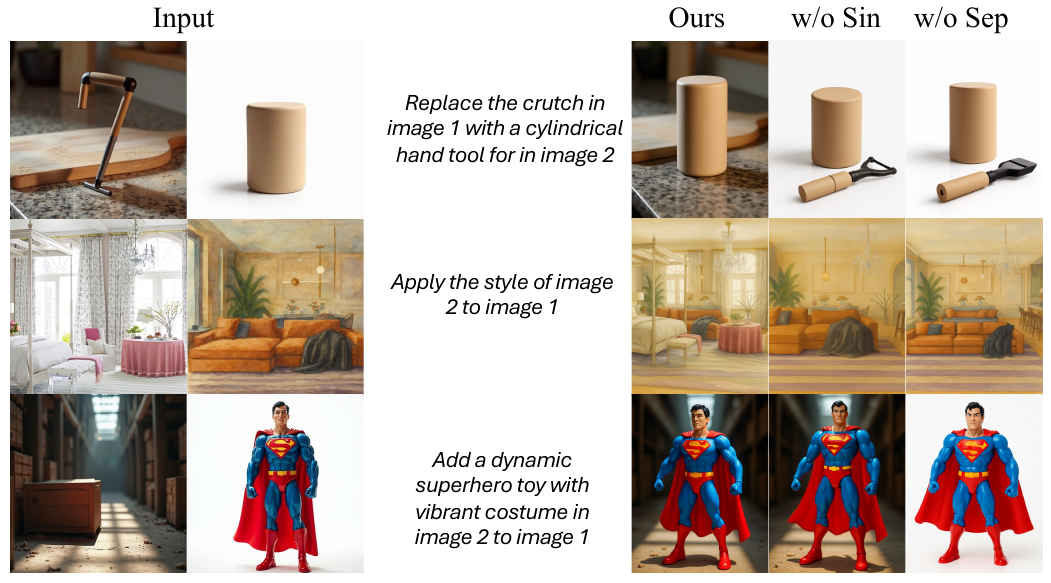}
\vspace{-7mm}
\caption{\textbf{Qualitative results for ablation study}. Removing the component may cause failure to cross-image reference and editing.}
\label{fig:abla_quali}
\vspace{-3mm}
\end{figure}

\noindent\textbf{Effectiveness Analysis}. We ablate the contributions of Separator and Sinusoidal Index Embedding in Table~\ref{tab:abl}. The combination of two modules achieves the best performance on most tasks. On the Style and Human tasks, although the combined module is 0.01 or 0.02 slightly weaker than the individual modules, the three modules get very close performance. Removing each module improves local specialization on a few cases, but the full version improves global generalization. Concretely, w/o Sinindex causes a clear 0.31 drop on Replace and 0.18 drop on Mixed. w/o Separator causes a 0.24 drop on Replace and 0.08 drop on Mixed, which reflects trade-offs between specialization and generalization. Thus, our model is generally effective in improving the multi-image editing. We also show qualitative results of ablation in Figure~\ref{fig:abla_quali}. The whole model can better correspond to the image identity and achieve the desired multi-image editing.

\begin{figure}[t]
\centering
\includegraphics[width=1.0\linewidth]{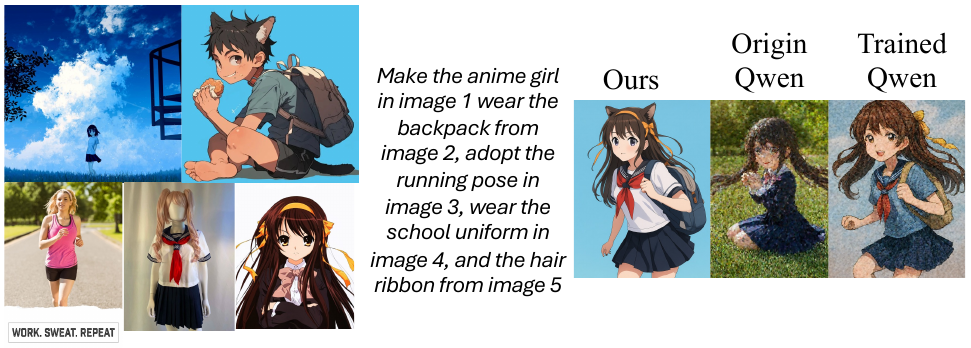}
\vspace{-7mm}
\caption{\textbf{Generalizaton evaluation}. We compare our model with the original Qwen-Edit-2509 and one trained on the same dataset. The training data does not include the 5-image input. Our model generalizes better to the extrapolated number of images.}
\label{fig:generalize}
\vspace{-3mm}
\end{figure}

\noindent\textbf{Generalization}. We evaluate the model’s generalization capability by testing it on a larger number of input images that \textit{never} appear during training. Specifically, we train our model primarily on two-image data and never on five-image inputs. For evaluation, we show the generalization test in Figure~\ref{fig:generalize}. The results show that compared to the original Qwen-Edit-2509, our model can produce the results aligned to the multimodal instruction, while the original Qwen-Edit-2509 does not generalize to the unseen 5-image set and produces noisy images. The Qwen-Edit trained on the same dataset shows better-aligned semantics but still tends to produce noisy images. This validates our generalization ability.

\noindent\textbf{Efficiency}. Despite introducing additional positional encodings and separator tokens, our model exhibits negligible differences in inference time compared to the original Qwen-Edit-2509. Both require approximately four minutes to complete a two-image editing task. Throughout all experiments, we use a separator token of size [1,1,64], which adds only a small number of extra tokens during inference and therefore incurs minimal computational overhead.

\begin{table}[t]
\centering
\tabcolsep=0.12cm
\caption{
\textbf{Ablation on MMIE-Bench evaluated by Qwen2.5-VL (72B)}. Learnable Separator Token (Separator), Sinusoidal Index Embedding (Sinindex), w/o indicates removing the module.
}
\vspace{-3mm}
\label{tab:abl}
\resizebox{\linewidth}{!}{
\begin{tabular}{lccccccc}
\toprule
 & \multicolumn{7}{c}{\textbf{Qwen2.5-VL Evaluation}} \\
\cmidrule(lr){2-8}
Method & Add & Replace & Style & Human & Reason & Mixed & Avg \\
\midrule
Qwen-Edit      & 2.99 & 3.00 & 2.56 & 2.72 & 2.75 & 2.67 & 2.77 \\
w/o Sinindex    & 3.68 & 3.20 & \textbf{3.12} & \textbf{3.23} & 3.12 & 3.12 & 3.26 \\
w/o Separator       & 3.72 & 3.27 & 3.11 & 3.23 & \textbf{3.23} & 3.22 & 3.29 \\
\textbf{Ours}  & \textbf{3.77} & \textbf{3.51} & 3.09 & 3.22 & 3.12 & \textbf{3.30} & \textbf{3.34} \\
\bottomrule
\end{tabular}
}
\vspace{-3mm}
\end{table}

\section{Conclusion}

We present a scalable multi-image editing framework for unified multimodal models (UMMs) that explicitly models the image indexes for better cross-image reference and visual consistency. 
Our design introduces two complementary algorithmic components: the learnable latent separator for explicit image-wise disentanglement and the generalized sinusoidal index encoding for continuous and extrapolative positional modeling across variable image counts. Together, these innovations enable the model to maintain coherent visual conditioning, resolve identity ambiguity, and generalize seamlessly to unseen multi-image configurations.
To support robust training and evaluation, we further established a high-fidelity benchmark through an inverse data construction methodology that aims for artifact-free and semantically grounded supervision. 
Comprehensive experiments on our MMIE-Bench validate the improvement in visual fidelity and consistency for multi-image editing.

{
    \small
    \bibliographystyle{ieeenat_fullname}
    \bibliography{main}
}

\clearpage

\clearpage
\setcounter{page}{1}
\maketitlesupplementary

\noindent The supplementary material is organized as follows:
\begin{itemize}
    \item Additional Related Work: 1) Instruction-based image editing, 2) Image editing dataset creation.
    \item Multi-image Editing Dataset Creation and MMIE-Benchmark.
    \item Additional Experiments and Analysis. 
    \item Limitations.
\end{itemize}

\section{Additional Related Work}
\noindent\textbf{Instruction-based Image Editing.} Instruction-based image editing methods generally adapt text-to-image models for editing tasks by fine-tuning them on triplets comprising source images, target images, and corresponding editing instructions. Broadly, these methods fall into three categories based on their generation paradigms: diffusion-based, autoregressive-based, and hybrid approaches. Currently, diffusion-based methods demonstrate superior image fidelity and flexibility compared to autoregressive counterparts. Specifically, InstructPix2Pix~\cite{brooks2023instructpix2pix} pioneered this direction by training a Stable Diffusion backbone~\cite{rombach2022high} with generated editing triplets. Subsequent studies have adopted similar strategies while incorporating more advanced text encoders, such as T5~\cite{vlt5} or Multimodal Large Language Models (MLLMs), to enhance multimodal understanding and support complex editing tasks~\cite{huang2024smartedit,liu2025step1x,qwen,wang2025seededit,labs2025flux_kontext,han2024ace,mao2025ace++}. Conversely, autoregressive-based methods generate visual tokens for the edited image sequentially. For instance, EditAR~\cite{mu2025editar} adapts LLamaGen~\cite{sun2024autoregressive} for editing by incorporating an additional CLIP alignment loss. Furthermore, recent approaches integrate autoregressive and diffusion models within a unified network. These native multimodal models aim to improve the synergy between visual and linguistic modalities~\cite{deng2025bagel,transfusion,seed-x}. However, most existing methods focus primarily on single-image editing and struggle to maintain visual consistency in multi-image contexts. Recent works such as Omnigen2~\cite{wu2025omnigen2}, DreamOmini2~\cite{xia2025dreamomni2}, and Query-Kontext~\cite{song2025query} address the multi-image setting by introducing a shift in Rotary Positional Embeddings (RoPE) to increase the \textit{relative} distance between images. In contrast, our method re-examines the arrangement of visual tokens within the MM-DiT architecture, explicitly adding image-wise separation and extrapolable index awareness.

\noindent\textbf{Image Editing Data Creation.} Generally, training image editing models necessitates datasets consisting of triplets: source images, target images, and editing instructions. Two factors are critical for data quality: the visual consistency between source and target images, and the semantic alignment between the instruction and the visual changes. To construct high-quality training data, several methods leverage existing atomic editing models~\cite{brooks2023instructpix2pix} or generation frameworks~\cite{hui2024hqedit,wu2025uso,wu2025uno_flux} to synthesize triplets. To enhance editing precision, UltraEdit~\cite{zhao2024ultraedit} incorporates object masks during generation. ShareGPT-4o-Image~\cite{chen2025sharegpt} utilizes the state-of-the-art GPT-4o~\cite{gpt4} to generate high-fidelity images. Similarly, Echo-4o~\cite{ye2025echo}, Pico-Banana~\cite{qian2025pico}, and GPT1.5m~\cite{wang2507gpt} employ advanced commercial generative models to synthesize editing data across diverse scenarios. While most approaches focus on single-image editing, Omnigen2~\cite{wu2025omnigen2} and Query-Kontext~\cite{song2025query} extend to multi-image settings by employing Grounding DINO~\cite{grounding_dino}, SAM~\cite{kirillov2023segment}, and inpainting models to extract and manipulate multiple objects. However, such pipeline approaches often introduce copy-paste artifacts, compromising editing fidelity. In contrast, our data generation pipeline synthesizes the same object across different scenes and perspectives, thereby avoiding such artifacts and ensuring natural coherence.

\section{Multi-image Editing Dataset Creation and MMIE-Benchmark}
\subsection{Dataset Construction}
\begin{figure*}[t]
\centering
\includegraphics[width=0.9\linewidth]{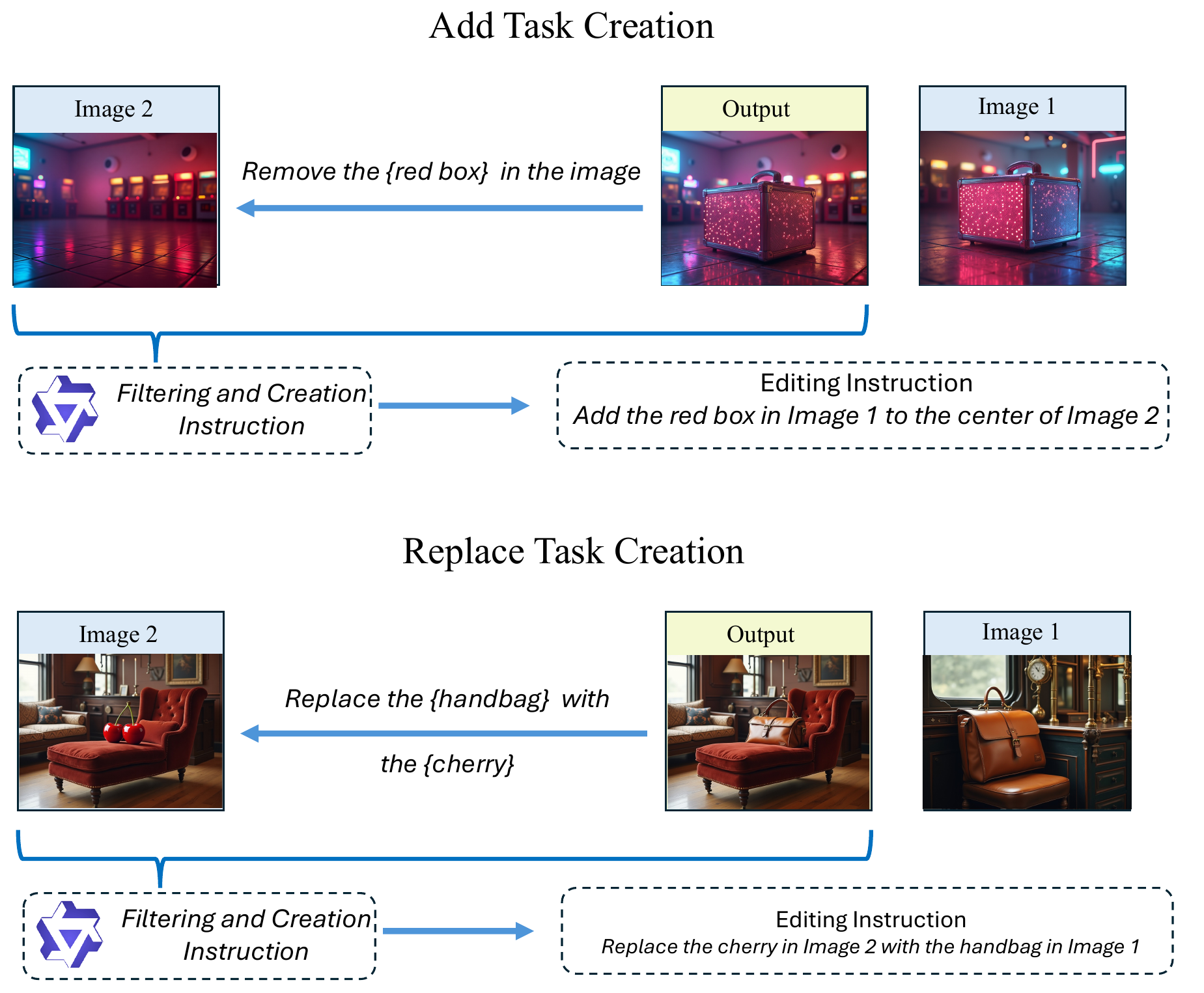}

\caption{\textbf{Dataset construction pipeline}. We show the data construction pipeline of the Add and Replace tasks. Initially, the primary object is identified in both images. For the Add task, we use the single-image editing model to remove the main object to get Image 2. For the Replace task, we randomly sample an object name from the LVIS name list and construct the editing instruction (e.g., replace the handbag with the cherry). Then we also use the single-image editing model to get Image 2. The post-filtering and processing are finally conducted on the edited images.}
\label{fig:data_pipe}
\end{figure*}

We present the detailed data construction pipeline in Figure~\ref{fig:data_pipe}. The construction is based on the Subject200k and UNO1M. For filtering the editing data, we use the Qwen2.5-VL to check the quality of the editing data. When removing the object, we check if the target object has been successfully removed or replaced, and if the background is naturally filled. When replacing the object, we check if the target object has been completely and perfectly replaced, and if the new object is free of deformities. These aim to filter images that have high fidelity and are well aligned with text instructions. The text prompt used for filtering is shown in the \textit{Filter Prompt}. The \{\} is filled with different objects according to the images. To cover as many objects as possible, we adopt the name list of LVIS~\cite{gupta2019lvis}, which comprises 1,200 objects commonly found in daily life.

\begin{tcolorbox}
\textit{Filter Prompt}:

I want you to help compare and analyze two images. 
You should check two things. 
First, compared with image 1, is \{\} shown in image 1 completely removed in image 2? 

Second, is the region of \{\} is recovered by the background in image2? 
If both are true, you should answer 'yes', otherwise, you should answer 'no'. 
Your answer should only include yes or no.
\end{tcolorbox}

\begin{figure*}[t]
\centering
\includegraphics[width=1\linewidth]{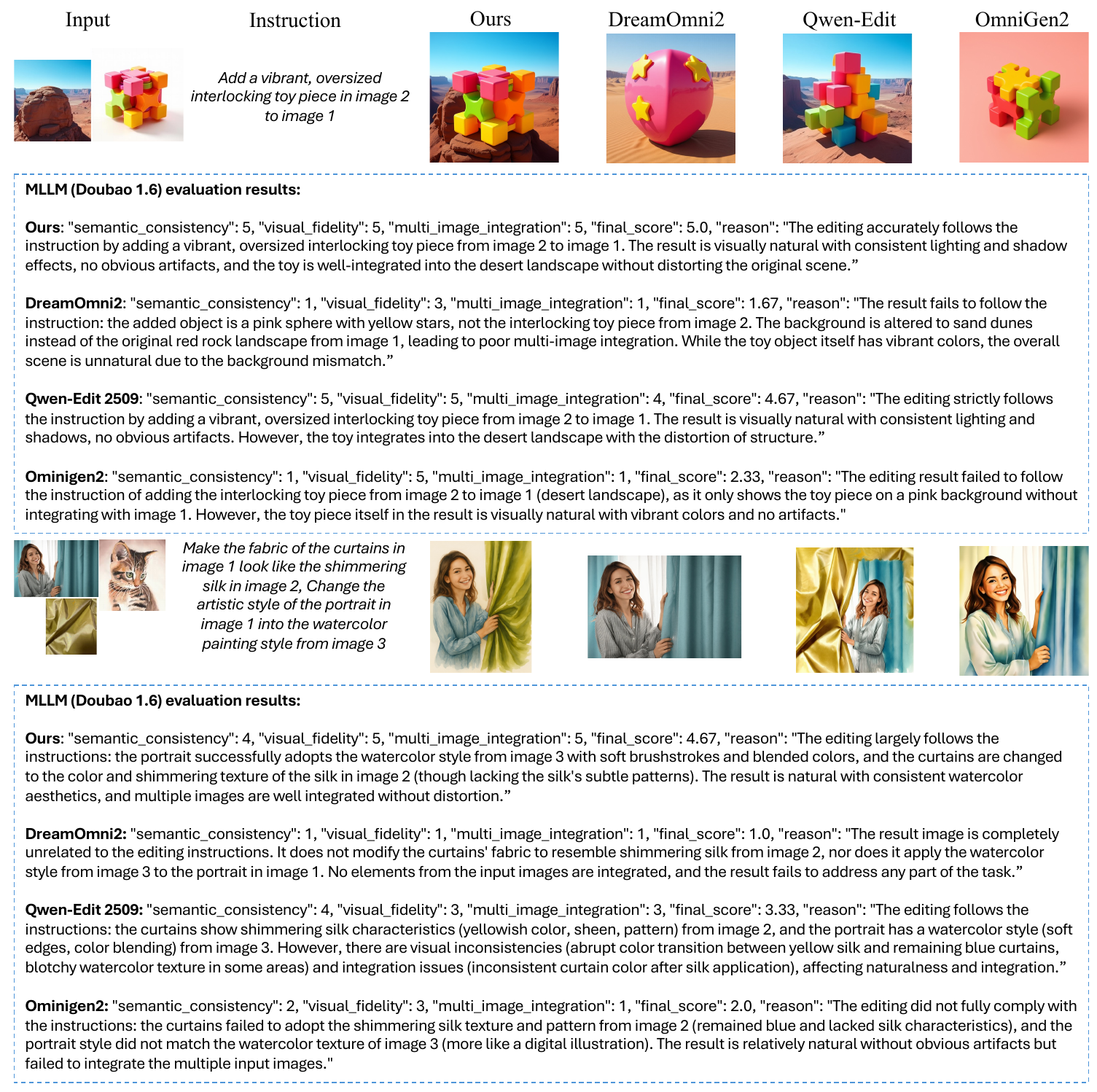}

\caption{\textbf{Detailed demonstration of MLLM evaluation results}. We show the detailed evaluation scores and corresponding reasons from Doubao-1.6 for different editing scenarios. The detailed reasons show that MLLM can overall evaluate correctly based on the given evaluation prompt. All human data is from Echo-4o.}
\label{fig:mllm_results}
\end{figure*}

\begin{figure*}[t]
\centering
\includegraphics[width=0.95\linewidth]{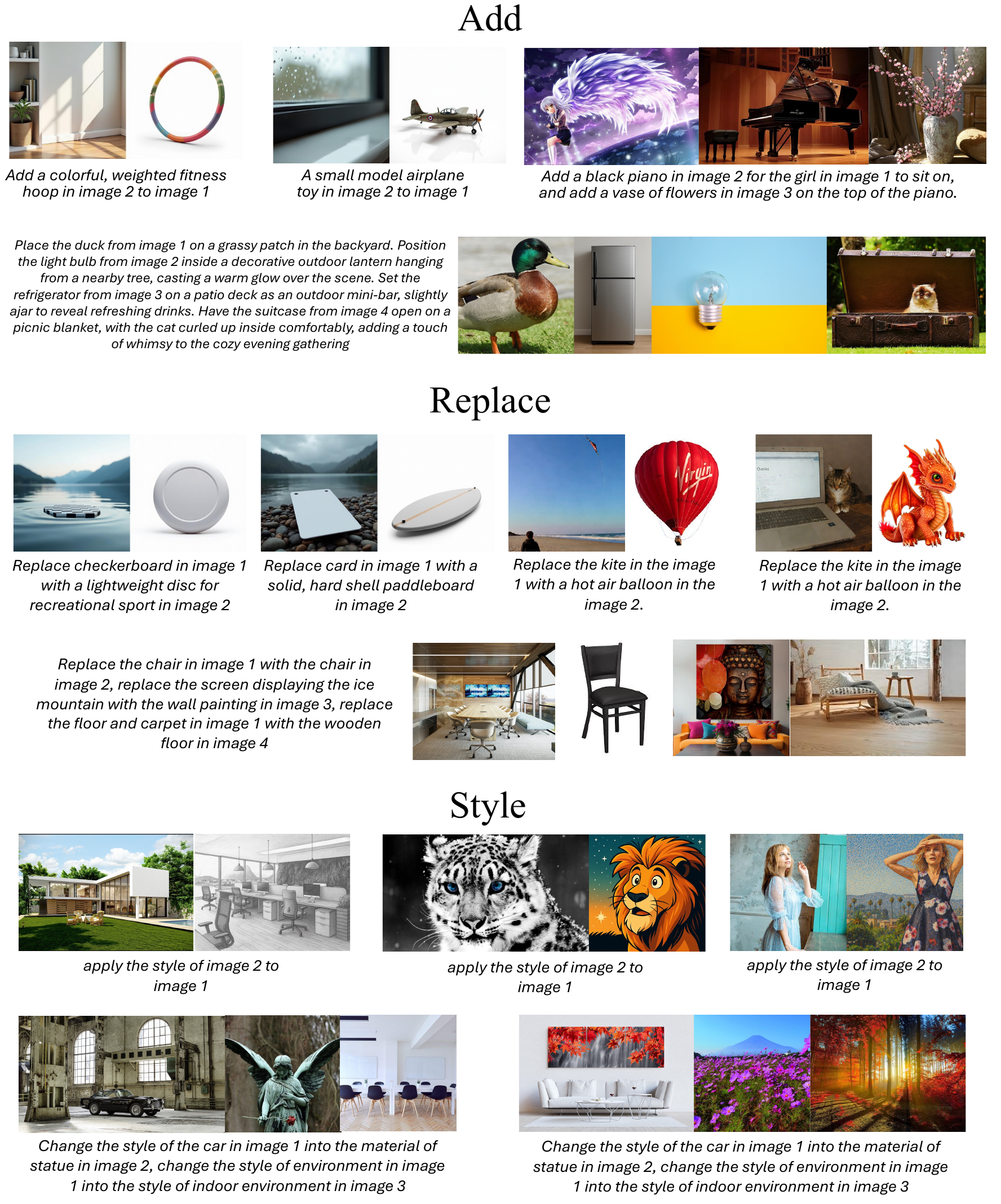}

\caption{\textbf{Demonstration of MMIE-Benchmark Part I}. All human data is from Echo-4o and PIE.}
\label{fig:bench1}
\end{figure*}

\begin{figure*}[t]
\centering
\includegraphics[width=0.95\linewidth]{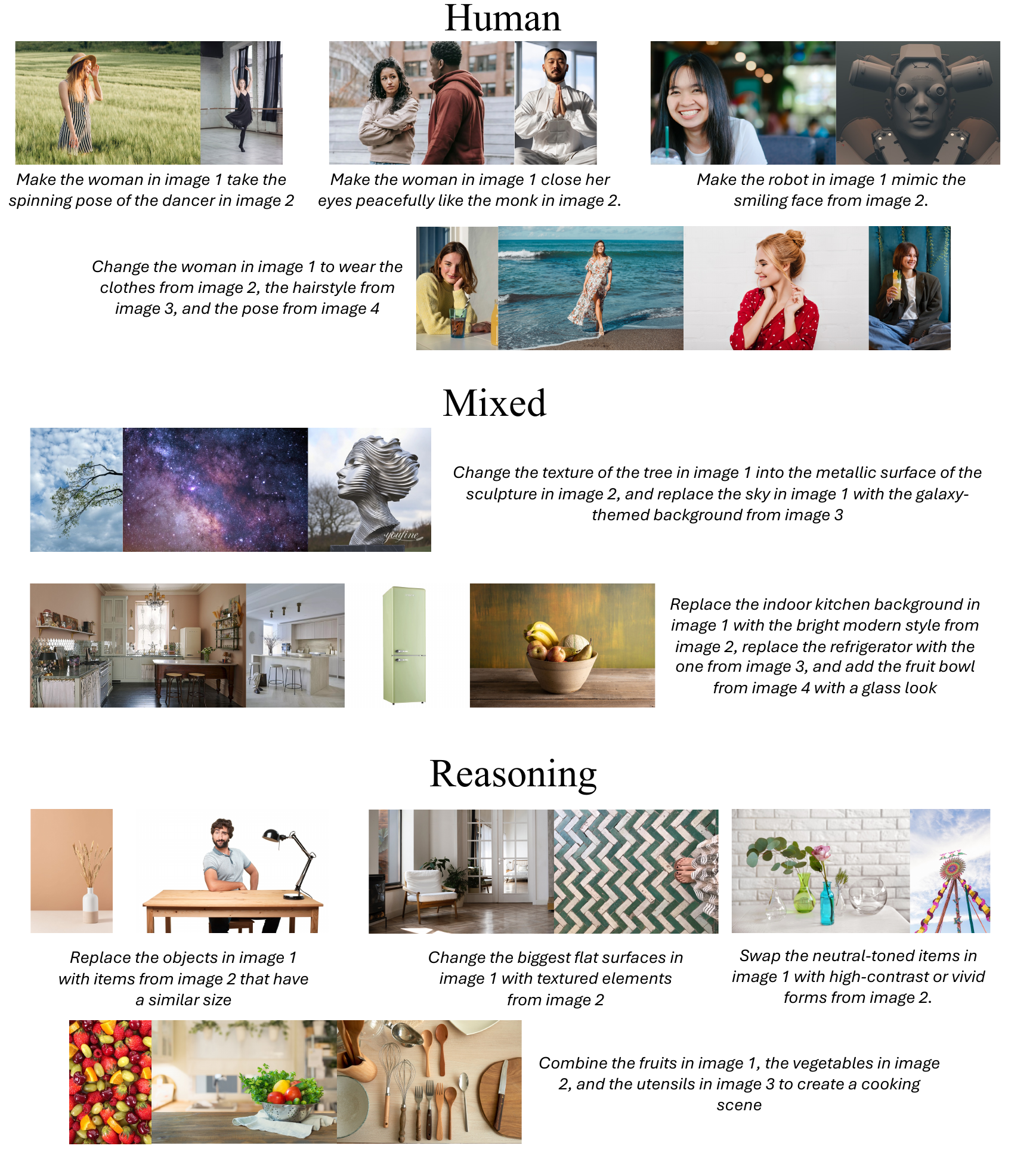}

\caption{\textbf{Demonstration of MMIE-Benchmark Part II}. All human data is from Echo-4o and PIE.}
\label{fig:bench2}
\end{figure*}

\subsection{MMIE-Benchmark}
\label{supp:bench}
We show the detailed examples of our benchmark in Figure~\ref{fig:bench1} and \ref{fig:bench2}. Our benchmark provides a comprehensive evaluation of \textit{different editing types, scenarios, numbers of input images, levels of geometric changes, reasoning abilities, and unifying generation and editing}. These considerations aim to comprehensively evaluate the visual and semantic consistency in multi-image editing and generation, which satisfies our motivation for UMMs. We also present our evaluation prompt in \textit{Evaluation Prompt}. We also report the number of input images among the evaluation cases in Table~\ref{tab:input_num}. We evaluate the performance on different numbers of input images in descending order, since the most common editing task involves two or three images.

\begin{tcolorbox}
\textit{Evaluation Prompt}:

You are an expert in image editing assessment.  
Please rate the input image, editing instructions, and result image based on the following three dimensions:
\\

1. semantic\_consistency: Whether the result correctly follow the editing instruction (1--5)\\
2. visual\_fidelity: Whether the result is natural and artifact-free (1--5)\\
3. multi\_image\_integration: Whether multiple input images are reasonably integrated without distortion (1--5)
\\

Please output JSON format, for example:

\{
  "semantic\_consistency": 4,\\
  "visual\_fidelity": 5,\\
  "multi\_image\_integration": 4,\\
  "final\_score": 4.33,\\
  "reason": "The editing complies with the instructions, the details are natural, and the multiple images are well integrated."
\}
\end{tcolorbox}

We show the detailed evaluation score and corresponding reasons from the MLLM in Figure~\ref{fig:mllm_results}. The detailed contents show that the MLLM can provide reasonable judgment based on the given prompt and three metrics. Concretely, the MLLM can distinguish the visual and semantic concepts of each image and indicate the obvious failures, such as background mismatch, failure to follow instructions, and integration distortion. However, it is also noted that the MLLM fails to detect some visual artifacts and differences. For example, in the 1st case, the Qwen-Edit produces a distorted interlocking toy, which does not decrease the VF score. In the 2nd case, the OmniGen2 produces a portrait whose clothes have been changed unexpectedly. This obvious error is not detected by the MLLM, while the MLLM claims the watercolor texture of the edited image does not match the texture of the reference cat, which is not as obvious as the difference in the clothes.

\begin{table}[t]
    \centering
    \caption{\textbf{Evaluation number of different input-image number in MMIE-Bechmark}.}
    \begin{tabular}{cccccc}
        \toprule
        Images      & 2 & 3 & 4 & 5 & Total \\
        \midrule
        Number         & 114 & 91 & 62 & 7 &274 \\
        \bottomrule
    \end{tabular}
\label{tab:input_num}
\end{table}

\begin{figure*}[t]
\centering
\includegraphics[width=0.95\linewidth]{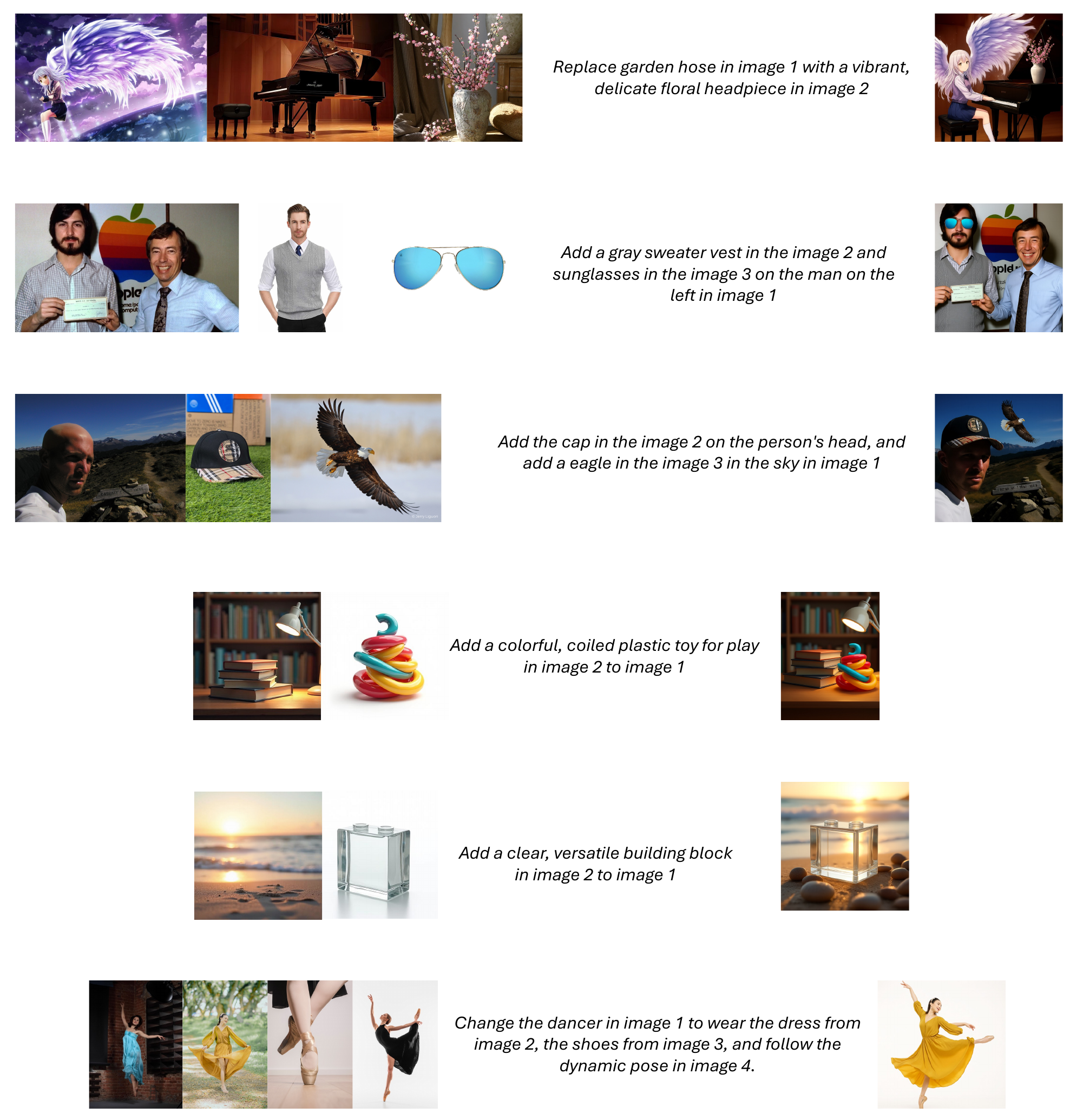}

\caption{\textbf{Additional qualitative results Part I}. All human data is from Echo-4o and PIE.}
\label{fig:add_qua1}
\end{figure*}

\begin{figure*}[t]
\centering
\includegraphics[width=0.95\linewidth]{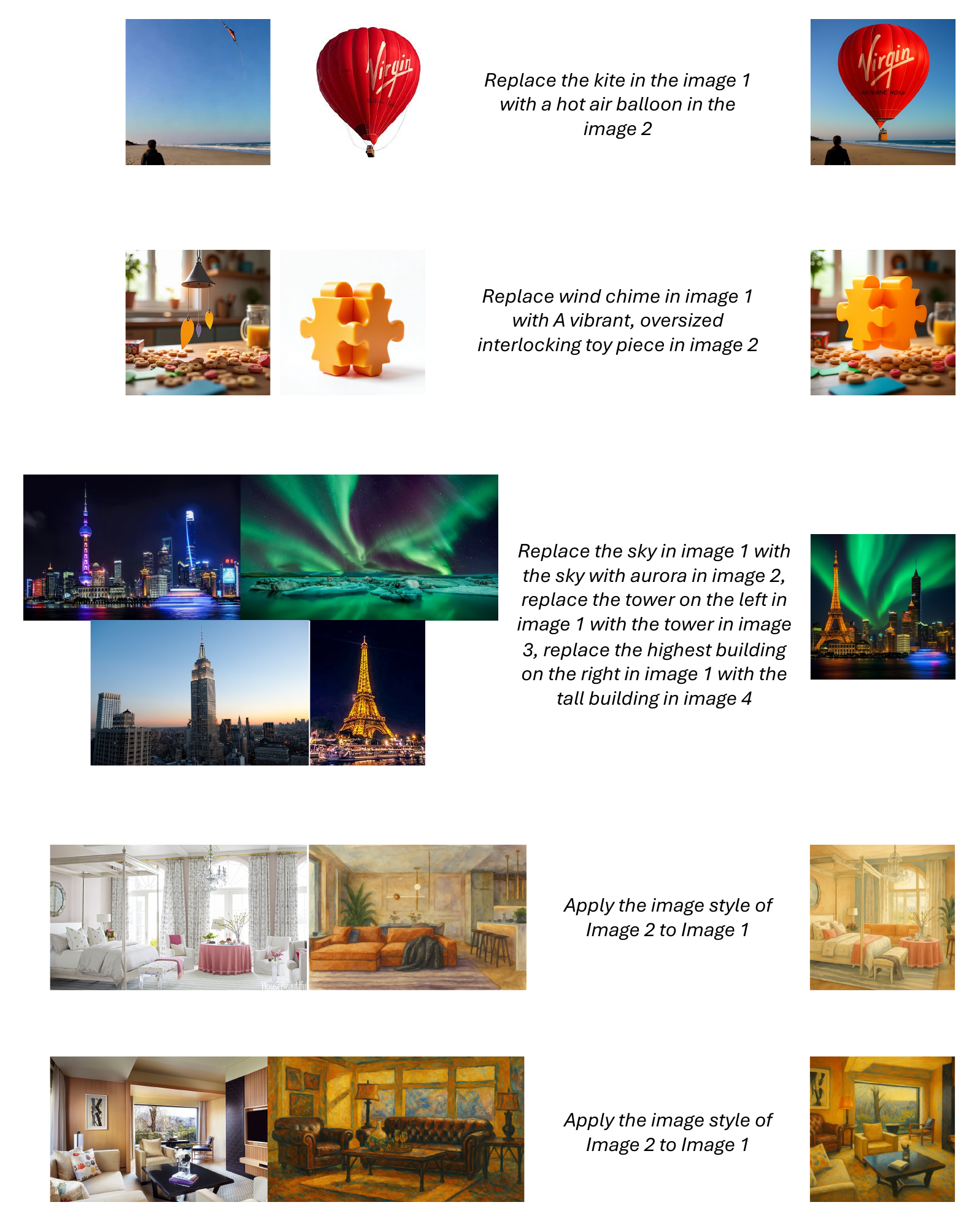}

\caption{\textbf{Additional qualitative results Part II}. All human data is from Echo-4o and PIE.}
\label{fig:add_qua2}
\end{figure*}

\section{Additional Experiments and Analysis}

\subsection{Ablation of Learnable Visual Separator Token}
To validate if the learning of the visual separator token is effective, we assign fixed values to the visual separator and only train the model, but not the separator itself. The separator is initialized with random Gaussian values and is fixed during training and inference. We compare the results in Table~\ref{tab:add_sep}. The results show that without the learning process, the performance clearly drops compared to the full model, which validates the effectiveness of the learning process.

\begin{table}[t]
\centering
\tabcolsep=0.12cm
\caption{
\textbf{Ablation on MMIE-Bench evaluated by Qwen2.5-VL (72B)}. Randomly initialized visual separator (Rand sep).
}
\vspace{-3mm}
\label{tab:add_sep}
\resizebox{\linewidth}{!}{
\begin{tabular}{lccccccc}
\toprule
 & \multicolumn{7}{c}{\textbf{Qwen2.5-VL Evaluation}} \\
\cmidrule(lr){2-8}
Method & Add & Replace & Style & Human & Reason & Mixed & Avg \\
\midrule
Rand sep      & 3.19 & 3.36 & 2.89 & 3.09 & 2.92 & 3.14 & 3.09 \\
\textbf{Ours}  & \textbf{3.77} & \textbf{3.51} & \textbf{3.09} & \textbf{3.22} & \textbf{3.12} & \textbf{3.30} & \textbf{3.34} \\
\bottomrule
\end{tabular}
}
\vspace{-3mm}
\end{table}

\subsection{Details of User Study}
\label{supp:user}
\begin{figure*}[t]
\centering
\includegraphics[width=0.9\linewidth]{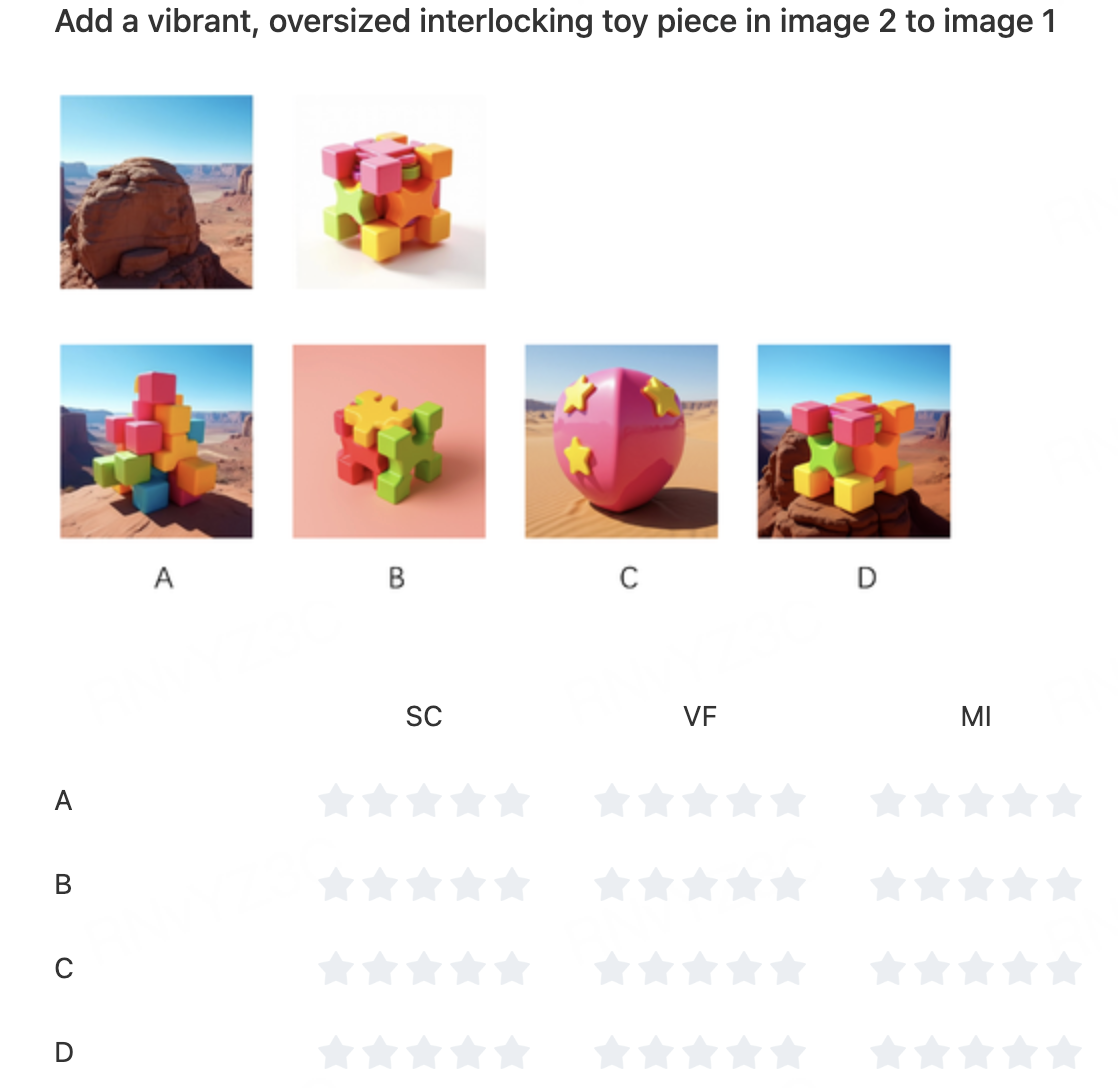}

\caption{\textbf{Interface of the user study}. The user evaluates each editing result according to three metrics (i.e., SC, VF, and MI) based on the input images and editing instructions.}
\label{fig:user_study}
\end{figure*}

The user study aims to evaluate the human preference for the three metrics of semantic consistency (SC), visual fidelity (VF), and multi-image integration (MI). To ensure fairness and consistency with the MLLM evaluation, each user is asked to evaluate the editing result using the same evaluation prompt shown in Section~\ref{supp:bench}. For each sub-task in MMIE-Bench, we randomly sample 5 cases for human evaluation. For each case, the user is asked to rate each method based on three metrics (SC, VF, and MI) from 1 to 5. The interface of the user study is shown in Figure~\ref{fig:user_study}. Each user is asked to evaluate 30 cases for 6 tasks in MMIE-Bench. We collected 23 users' answers to calculate the average score for each editing method, which is the same as the MLLM evaluation.

\subsection{Additional Qualitative Results} 
We show additional qualitative results in Figures~\ref{fig:add_qua1} and \ref{fig:add_qua2}. Our method understands multiple images and text instructions, and generates the image with both visual and semantic alignment in different scenarios, editing types, and object types. These qualitative results validate the effectiveness and robustness of the multi-image editing.

\section{Limitations}
We discuss the limitations from two perspectives: the model design and the evaluation. For the module design, theoretically, the proposed sinusoidal index embedding can represent and extrapolate the index of many images, but it may not generalize to a very large number of input images. This is because the periodic property of the sinusoidal function may make the index embedding ambiguous. On the other hand, the input images may not go up to such a large number. Most editing involves images of fewer than ten. For the evaluation, the current evaluation is bounded by the ability of the MLLM. Some visual content, such as hand distortion and fine-grained details, is not effectively evaluated by the MLLM and reflected in the evaluation score. Using stronger and better MLLM can get more precise evaluation results.

\end{document}